\documentclass[sigconf]{acmart}
\pdfoutput=1
\usepackage{xspace}
\usepackage{graphicx}
\usepackage{amsmath}
\usepackage{booktabs}
\usepackage{enumitem}
\usepackage{multirow}
\usepackage{appendix}
\usepackage{graphicx} 
\usepackage{subfigure}
\usepackage{amsmath}

\AtBeginDocument{%
  \providecommand\BibTeX{{%
    \normalfont B\kern-0.5em{\scshape i\kern-0.25em b}\kern-0.8em\TeX}}}

\renewcommand\footnotetextcopyrightpermission[1]{}
\begin{document}
\fancyhead{}

\def\eg{\emph{e.g.,}} \def\Eg{\emph{E.g.,}}          
\def\ie{\emph{i.e.,}} \def\Ie{\emph{I.e.,}}          
\def\cf{\emph{c.f.}} \def\Cf{\emph{C.f.}}          
\def\etc{\emph{etc.}}                              
\def\vs{\emph{vs.}}                                
\def\etal{\emph{et al.}}                           
\newcommand{\onestage}{\emph{One-Stage Pipeline\xspace}}
\newcommand{\videopc}{\emph{Video Paragraph Captioning\xspace}}

\newcommand{\Hline}{\Xhline{2\arrayrulewidth}}     
\newcommand{\HLINE}{\Xhline{4\arrayrulewidth}}     
\newcommand{\dash}{-\phantom{00}}                  
\newcommand{\tstrut}{\rule{0pt}{2.0ex}}            
\newcommand{\bstrut}{\rule[-0.9ex]{0pt}{0pt}}      
\newcommand{\tcite}[1]{\scriptsize{\cite{#1}}}     
\newcommand{\tcitep}[1]{\scriptsize{\citep{#1}}}   

\title{CapOnImage: Context-driven Dense-Captioning on Image}

\author{Yiqi Gao$^{1}$\footnotemark[1] \quad Xinglin Hou$^{2}$ \quad Yuanmeng Zhang$^{2}$ \quad Tiezheng Ge$^{2}$ 
\quad Yuning Jiang$^2$ \quad Peng Wang$^{1}$} 

\affiliation{
 \institution{\textsuperscript{\rm 1}School of Computer Science, Northwestern Polytechnical University} 
 \institution{\textsuperscript{\rm 2}Alibaba Group}
 \country{}
 }
 
\email{gyqjz@mail.nwpu.edu.cn, peng.wang@nwpu.edu.cn}
\email{{xinglin.hxl, zhangyuanmeng.zym, tiezheng.gtz, mengzhu.jyn}@alibaba-inc.com}

\def\authors{Yiqi Gao, Xinglin Hou, Yuanmeng Zhang, Tiezheng Ge, Yuning Jiang and Peng Wang}
\renewcommand{\shortauthors}{Yiqi Gao, et al.}

\begin{abstract}
	Existing image captioning systems are dedicated to generating narrative captions for images, which are spatially detached from
	the image in presentation.
	However, texts can also be used as decorations on the image to highlight the key points and increase the attractiveness of images.
	In this work, we introduce a new task called captioning on image (CapOnImage), which aims to generate dense captions at different locations of the image based on contextual information.
	To fully exploit the surrounding visual context to generate the most suitable caption for each location, we propose a multi-modal pre-training model with multi-level pre-training tasks that progressively learn the correspondence between texts and image locations from easy to difficult. 
	Since the model may generate redundant captions for nearby locations, we further enhance the location embedding with neighbor locations as context.
	For this new task, we also introduce a large-scale benchmark called CapOnImage2M, which
	contains 2.1 million product images, each with an average of 4.8 spatially localized captions.
	Compared with other image captioning model variants, our model achieves the
	best results in
	both captioning accuracy and diversity aspects. We will make code and datasets public to facilitate future research.
\end{abstract}

\maketitle

\renewcommand{\thefootnote}{\fnsymbol{footnote}}
\footnotetext[1]{Work done during an internship at Alibaba Group}

\section{Introduction}

\begin{figure}
  \centering
  \includegraphics[width=0.95\linewidth]{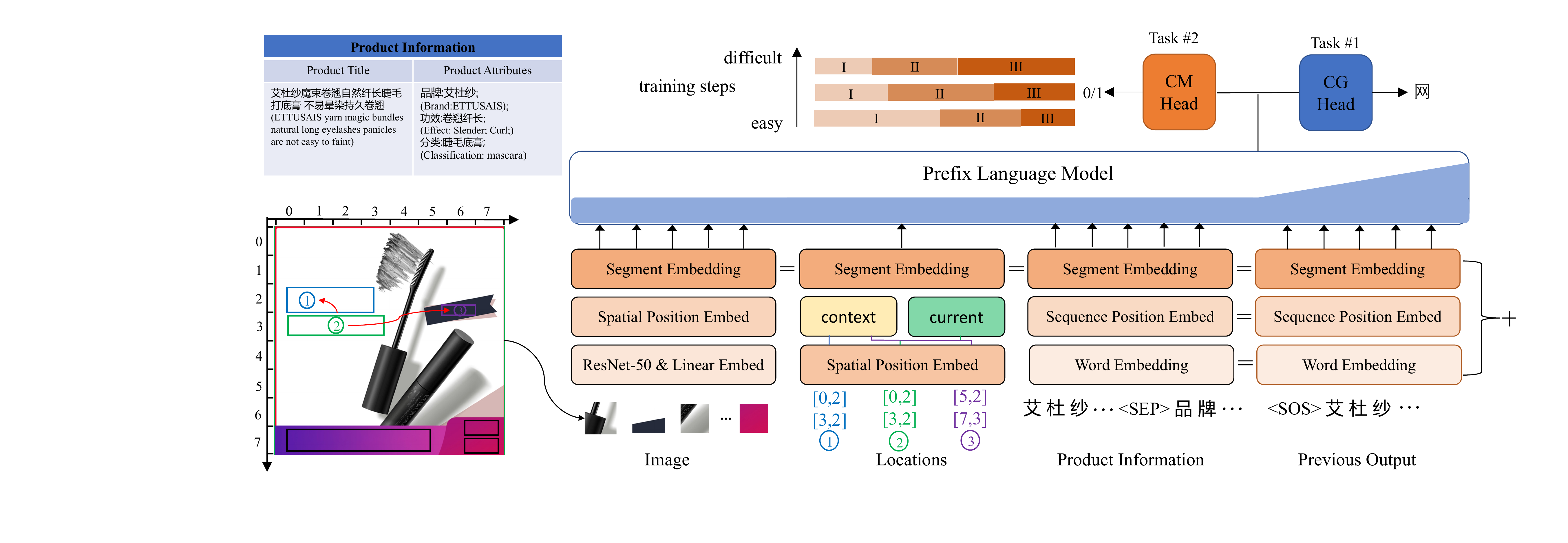}
  \caption{Illustration of descriptive texts on image scenarios. The captions and 
  images in existing tasks (\eg visual genome (a) \cite{krishna2017vg}) are spatially detached, without any association between each other.
  In contrast, our CapOnImage task aims to generate descriptive text \emph{on} the image, which has strong necessity and broad application prospects in some scenarios ((b) and (c)).
  }
  \label{fig:example}
  \vspace{-8pt}
\end{figure}

Building upon the advances in computer vision and natural language processing areas, the new research direction called vision-and-language has attracted more and more attentions, which pushes to tackle new problems that need to bridge the two areas to advance the concept comprehension and reasoning capabilities. The image captioning task, as one of the most classic vision-and-language tasks, aims to generate natural language descriptions/captions for images \cite{vinyals15showtell,anderson2018butd,zhang21rstnet,johnson16densecap}.

However, the captions and the images in this task are spatially detached in presentation, without any association between each other (Figure~\ref{fig:example}(a)).  
In fact, there are many scenarios where the image and text are tightly associated.
For example, the product images on e-commercial website (Figure~\ref{fig:example}(b)) usually contain descriptive texts
, explaining multiple perspectives of the product (e.g., product characteristics, selling points, etc)
, which makes the image more informative and attractive.
On the social media platform, users usually upload daily pictures with descriptive texts as decoration
 (Figure~\ref{fig:example}(c)).
Therefore, it is significant
to explore captioning \emph{\textbf{on}} the image, which requires to consider not only the visual description, but also the description placement on the image.
Besides, to generate the informative captions in these scenarios, additional textual knowledge is usually needed, such as the product information and the background story about the image etc.
Therefore, in this work, we introduce a new task called \emph{CapOnImage}, which aims to generate dense captions at different locations of the image based on contextual information.
The CapOnImage task involves two steps, where the model needs to first predict a reasonable and aesthetical text layout \cite{arroyo21layout, gupta20layout,jyothi2019layoutvae}, and then 
generates a phrase or sentence for each text box.
In this work, we simplify this task to generate captions for a provided list of text box locations, thereby removing the requirement for layout prediction.
Therefore, the main focus of this work is to generate captions that are most suitable for the corresponding image locations.

The CapOnImage task involves two new challenges: \emph{(i):} Better understanding of context information: the captions at different locations can be greatly diverse. 
As shown in Figure~\ref{fig:example}(b), the texts around the product are descriptive captions describing the product features, while those at the bottom introduce the selling points. 
Therefore, the model needs to fully exploit the visual context around the text box to determine what caption is suitable to generate here.
Since our task aims to generate captions \emph{on} image, location context is vital for our task which is also validated on Table~\ref{tab:sota_comparison}(no-locations). 
Overall, compared with traditional caption task, the CapOnImage task needs better understanding of context information.
\emph{(ii):} Caption redundancy: some texts can be suitable for adjacent locations.
Therefore, if the model can only ``see'' the current text box without surrounding ones, it tends to generate the same caption for nearby text boxes because it suits all of them, thus causing the problem of caption redundancy.

To solve the above mentioned challenges, we propose a multi-modal pre-training and fine-tuning framework which contains multi-level pre-training tasks to effectively exploit multi-modal context information. \textbf{First}, to better understand context information, we design multi-level pre-training tasks helping the model ``feel'' the context information. It explicitly equip the model with the ability to distinguish which captions are appropriate for the current location and image while which are not. Besides, inspired by the progression of easy to complex biological vision systems, we further propose a progressive training strategy which learns multi-level pre-training tasks from easy to difficult. \textbf{Second}, to solve the problem of caption redundancy, we introduce a neighbor-enhanced location encoding module, which utilizes the surrounding text box locations as context, so that our model can ``see'' the adjacent context.
We show the captioning diversity results with different ranges of adjacent text boxes involved in the location encoding module, and demonstrate the importance of such neighbor context.

In order to evaluate our model and benchmark progress in the CapOnImage task, we introduce the \textit{CapOnImage2M} dataset.
It contains 2.1 million product images crawled from an e-commercial website, and each image contains multiple spatially localized captions describing the associated product.
We automatically acquire the text contents and their spatial locations from the image via OCR \cite{li2017ocr,liu2018ocr}, and finally collect 4.8 captions for each image on average.
We also crawl the product title and attributes as additional context information for caption generation.
With the CapOnImage2M dataset, we provide the first result for the CapOnImage task.
We show that both the visual context, location information and the additional product information are necessary for the caption generation, and our model can generate different types of captions at different spatial locations (Figure~\ref{fig:type}).
Furthermore, we demonstrate that our proposed neighbor-enhanced location encoding module and multi-level pre-training tasks significantly improve the captioning accuracy and diversity.

The main contributions of this work are as follows:
\parskip=0.1em
\begin{itemize}[itemsep=0.5pt,partopsep=0pt,parsep=\parskip,topsep=0.5pt]
    \item We introduce a new vision-and-language task called CapOnImage, which requires the model to tightly associate image and texts as a whole.
    \item We analyze the challenges of CapOnImage task and propose a context enhanced model with progressive training strategy, which achieves the best result compared with other image captioning model variants.
    \item We propose a large-scale multi-modal dataset called CapOnImage2M, with 50 categories images and localized captions to support the CapOnImage research.
\end{itemize}
\section{Related Work}
In recent years, significant progress has been made in the image captioning task \cite{vinyals15showtell,anderson2018butd,huang2019aoanet,pan2020xlnet,zhang21rstnet}, which aims to describe the image content in one natural sentence.
With the advances in visual understanding abilities, researchers are not satisfied with generating dull and less informative captions and upgrade the traditional image captioning task along two directions: describing image in multiple sentences to cover more details, and describing with additional knowledge to make the caption more informative.

The first direction is called dense captioning \cite{johnson16densecap,Melas2018paragraph,krishna17densecap,wang2021pdvc,song2021paragraph}, which targets to describe detailed visual content with a set of sentences.
Johnson \etal ~\cite{johnson16densecap}
propose a fully convolutional localization network to unify the object detection \cite{sermanet2014obj,girshick2014obj,ren2015fasterrcnn} and image captioning in one framework to predict a set of descriptions across object regions.
Krishna \etal ~\cite{krishna17densecap} migrate it to the video, which aims to predict sequential event proposals and generate description for each clip.
In these works, the dense captions deliver more details of visual content than traditional single sentence.
The second direction is called text-aware image captioning \cite{biten2019goodnews,sidorov20textcap,yang2021tap}, where the model generates captions not only according to the image, but also utilizes additional textual information as context.
Works in \cite{ramisa2018breakingnews, biten2019goodnews} propose entity-aware captioning for news images with news articles as context.
The background knowledge from additional textual input helps the model generate more informative captions containing named entities.
Some other works \cite{lu2018emnlp,zhao2019acl} explore to use external web source as the knowledge.
Besides the entity-aware captioning, Sidorov \etal ~\cite{sidorov20textcap} and Gurari \etal ~\cite{gurari2020viswiz} propose to generate image captions with scene texts, which exploit OCR tokens as the textual context.

Although impressive progresses have been made along the two directions, they remain separate.
The proposed CapOnImage task can be considered as a combination of the two directions, where the model needs to first predict the text layout (spatial locations on the image) and then generate caption for each location conditioned on both the image and textual information.
Moreover, in conventional image captioning task and its variants, the generated captions are detached from the image as only textual narrations, while in the CapOnImage task, the model generates captions that tightly associated with the image as a whole.
The caption length and type also need to be suitable to the corresponding location, which are not considered in previous image captioning tasks.
The CapOnImage task also promotes the research of layout generation \cite{arroyo21layout, gupta20layout,jyothi2019layoutvae}, from the document layout design \cite{zhong2019document} and user interface design \cite{deka2017rico} to text on image layout design.

\section{CapOnImage2M Dataset}

In this section, we introduce our proposed CapOnImage2M dataset, which is the benchmark for the CapOnImage task.
We first present an overview of the dataset collection and statistics, and then compare it with other related image captioning datasets.
A more detailed datasheet describing the motivation, composition, and recommended uses of our CapOnImage2M dataset following \cite{gebru2018datasheet} can be found in the supplementary material.

\subsection{Dataset Collection and Statistics}
\begin{figure}
    \centering
    \includegraphics[width=\linewidth]{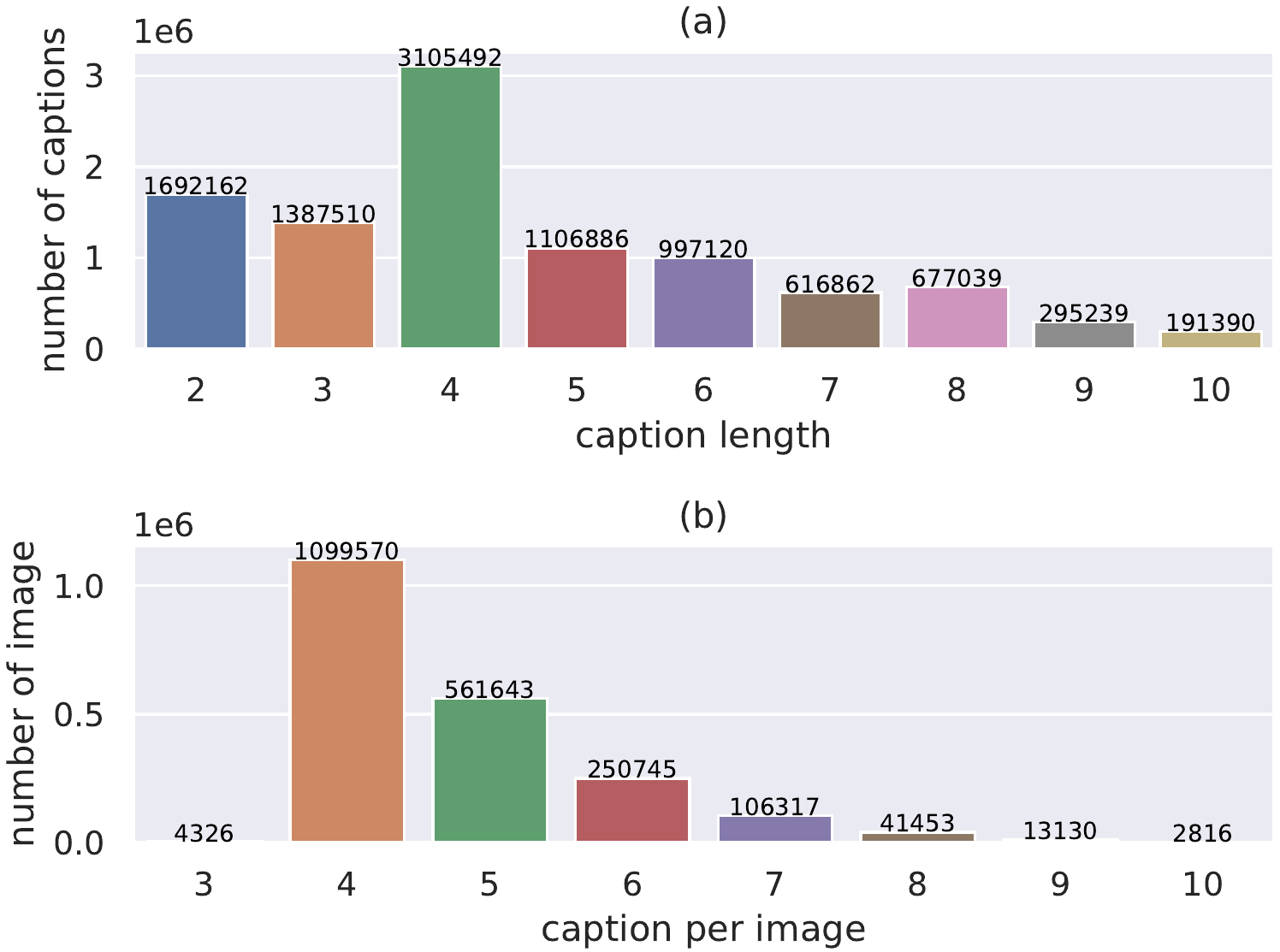}
    \vspace{-12pt}
    \caption{\emph{(a)}: Length distribution of caption. \emph{(b)}: Distribution of captions per image.}
    \vspace{-4pt}
    \label{fig:len}
\end{figure}

The CapOnImage2M dataset contains 2.1 million product images crawled from a Chinese e-commercial website\footnote{https://taobao.com}, where each image contains both the product and descriptive captions describing the product features, efficacy, brand and so on.
We crawl the images with 50 product categories that can cover most of the product(detail information, \eg word cloud, can be found in supplementary).
We also crawl the product title and attributes as the product information context.
For each image, we employ an OCR toolkit to recognize the texts and their spatial locations on the image.
We remove the texts longer than 10 or shorter than 2 characters, and remove redundant images with highly overlapped captions.
To further clean the noise in the texts, we input the auto-recognized captions into a pre-trained GPT \cite{radford2019language} model and remove those with high perplexities.
We also remove the texts that highly correlate with company marketing strategies, \eg discount information,  since it depends on the company and thus cannot be determined by an AI.
Finally, we acquire 2.1 million images, each with an average of 4.8 captions.
Each caption has an average length of 4.9 Chinese characters. We show the distribution of caption length and number of captions per image in Figure~\ref{fig:len}.
We split the dataset into 2.06M images for training, 20k for validation, and 20k for testing.
In order to ensure the accuracy of the evaluation, we further manually clean  and modify the OCR errors in the captions of the validation and test set through a public Chinese crowdsourcing platform.
During training and inference, we mask the pixels of all the texts on the image according to their bounding box coordinates, otherwise, the model will be degraded into an OCR model to predict text according to text pixels.
Examples of the dataset can be found in the supplementary material.

\begin{table}[t]
    \caption{Comparisons of different datasets. CI: captioning on image, IC: image captioning, FC: fashion captioning. }
    \label{tab:datasets}
    \centering
    \small
    \begin{tabular}{l c c c c c}
    \toprule
    dataset & \#image & \#text & avg\_len & dense & task \\
    \midrule
    CapOnImage2M & 2.1M & 10.07M & 4.9 & \checkmark & CI \\
    \midrule
    Flickr30K\cite{young2014flickr30k} & 30K & 150K & 12.3 & - & IC \\
    MSCOCO\cite{lin2014mscoco} & 123K & 616K & 10.4 & - & IC \\
    VG\cite{krishna2017vg} & 108K & 5M & 5.7 & \checkmark & IC \\
    TextCaps\cite{sidorov20textcap} & 28K & 142K & 12.4 & - & IC \\
    FACAD\cite{yang2020facad} & 993K & 130K & 21.0 & - & FC \\
    \bottomrule
    \end{tabular}
\end{table}

\subsection{Comparison with Other Datasets}
In Table~\ref{tab:datasets}, we compare our CapOnImage2M dataset with other image captioning datasets.
The CapOnImage2M dataset is substantially larger in both the number of images and texts.
Unlike VG\cite{krishna2017vg}, where the dense captions independently describe different regions of the image, the CapOnImage2M dataset contains dense captions that describe the same product from different aspects.
In addition to the dense captions on the image, each image also comes with a product title and attributes with an average length of 34.8 characters as the textual context in the CapOnImage2M dataset.
Therefore, it can also support the fashion captioning research as the FACAD\cite{yang2020facad} dataset does.

\section{Model}

\begin{figure*}
  \centering
  \includegraphics[width=\linewidth]{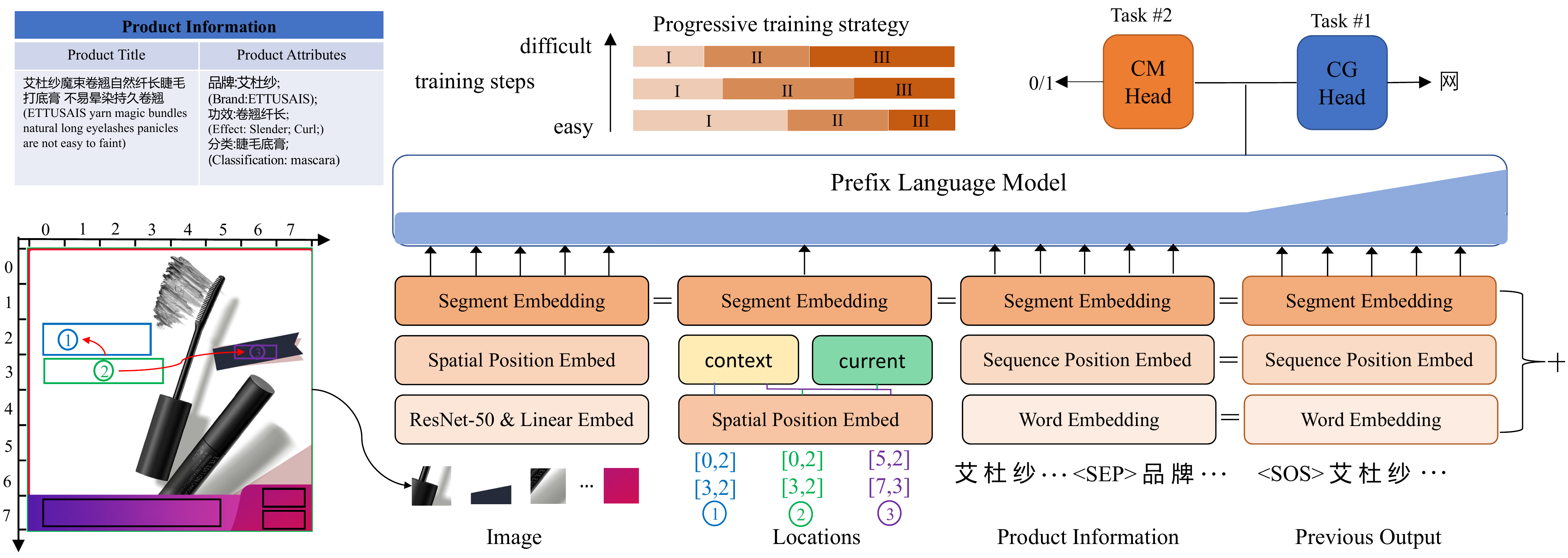}
  \caption{Illustration of our model with four input modalities: image patches, location coordinates, product information, and the predicted text tokens. Two pre-training tasks are employed to optimize the model with a progressive learning strategy from easy to difficult.``='' denotes parameter sharing and ``+'' denotes addition. We add English translation for product information for better understanding.}
  \vspace{-8pt}
  \label{fig:model}
\end{figure*}

In this section, we introduce our captioning on image model based on the pre-training and fine-tuning framework as illustrated in Figure~\ref{fig:model}.
Given an image, a location coordinate, and product information, our model aims to generate the most suitable caption for the current textbox location.
We extract feature representations for each modality respectively and then concatenate them into a multi-modal sequence.
A multi-layer transformer is applied on the multi-modal input to capture cross-modal contexts for caption generation.
To make the model aware of nearby text layout to avoid generating redundant captions, we enhance the location embedding with neighbor locations as context.
Moreover, a progressive training strategy with multi-level pre-training tasks is proposed to enhance the correspondence learning between textbox locations and captions.

\subsection{Input Representation}
The inputs of our model include three parts from different modalities: the visual image, the textbox location coordinate and the textual product information.
We independently encode each modality input as a sequence of $d$-dimensional feature vectors as follows.

\noindent\textbf{Image representation.}
Given the image, we extract the grid features with standard ResNet-50 \cite{he2016resnet} backbone, which is further end-to-end fine-tuned with our model.
We flatten the $k \times k$ feature map into a sequence and add spatial position embedding similarly as DETR \cite{carion2020detr}.
Specifically, for the $i$-th grid whose horizontal and vertical indexes are $x_i$ and $y_i$, we add learnable spatial embedding and segment embedding which indicates the image modality to the appearance feature $v_i$ as follows:
\begin{equation}
    \hat{v}_i = v_i + [Emb_h(x_i); Emb_v(y_i)] + \text{SE}_v,
\end{equation}
where $Emb_h(\cdot)$ and $Emb_v(\cdot)$ are horizontal and vertical embedding layers with the output dimension of $\frac{d}{2}$, $[;]$ denotes concatenation and SE denotes segment embedding.
Finally, we represent the image with a sequence of patch features $\hat{V}=\{\hat{v}_1,\cdots,\hat{v}_{k \times k}\}$.

\noindent\textbf{Neighbor-enhanced location representation.}
\label{sec:neighbor location}
We represent a text box location with 2D coordinates $\{(x_{min},y_{min}),(x_{max},y_{max})\}$, where $(x_{min},y_{min})$ is the top left corner coordinate and $(x_{max},y_{max})$ is the bottom right corner coordinate.
We map the real value coordinates into the $k \times k$ grid and represent them with the same spatial position embeddings as image:
\begin{equation}
    e_{cur} = [Emb_h(x_i); Emb_v(y_i); Emb_h(x_j); Emb_v(y_j)],
\end{equation}
where $\{(x_i, y_i),(x_j, y_j)\}$ is the corresponding grid index.

Furthermore, to avoid the problem of caption redundancy for adjacent locations, we enhance the location representation with neighbor locations as context.
We define the distance of two text boxes as the distance of their centers, the text boxes whose upper left corner are with smaller value of x-coordinate plus y-coordinate than the current one as the previous text boxes, and those larger than the current one as the next text boxes.
Then, we employ the nearest previous textbox location and the nearest next textbox location as the neighbor context, and encode them similarly as $e_{cur}$.
After encoding, we concatenate them with the current location embedding and add a segment embedding indicating the location modality as follows:
\begin{equation}
    l = [W_1^T \cdot [e_{prev}; e_{next}]; W_2^T \cdot e_{cur}] + \text{SE}_l,
\end{equation}
where $W_1 \in \mathbb{R}^{4d \times \frac{d}{2}}$ and $W_2 \in \mathbb{R}^{2d \times \frac{d}{2}}$
are learned matrices, $e_{prev}$ and $e_{next}$ are the neighbor location embeddings, SE is the segment embedding.

\noindent\textbf{Product information representation.}
To generate informative product descriptions, we also exploit product information as the textual context, which is the product title and attribute in this work. We concatenate them with a special $<SEP>$ token. 
Given the product information $X=\{x_1,\cdots,x_K\}$ with $K$ words, we embed these words via the same word embedding matrix as the target caption words, and add positional and segment embeddings as follows:
\begin{equation}
    w^{\text{info}}_i = W_e \cdot x_i + \text{PE}_i + \text{SE}_{x},
\end{equation}
where $W_e$ is the word embedding matrix, PE denotes sequence positional embedding as in BERT \cite{devlin2019bert} and SE denotes segment embedding.
Finally, we represent the product title with a sequence of $d$-dimensional feature vectors as $W^{\text{info}}=\{w_k^{\text{info}}\}^K_{k=1}$.

\subsection{Pre-training Tasks and Strategy}
After encoding each input modality into the common embedding space, we employ transformer layers on the multi-modal input to fuse the multi-modal information.
To generate appropriate and diverse descriptions at different textbox locations, we pre-train the model with two pre-training tasks, including Caption Generation (CG) and Caption Matching (CM).
We first pre-train the model with both CG and CM tasks, and then fine-tune it only with the CG task for the final caption generation.

\noindent\textbf{Task \#1: Caption Generation (CG).}
We generate captions using the same multi-modal transformer layers as decoder following the prefix LM \cite{raffel2020prefixLM, dong2019unilm}.
Each word prediction can attend to all the image features, neighbor-enhanced location and product information embeddings, as well as previous generated words.
We adopt the auto-regressive training objective for the CG task, which can be expressed as follows:
\begin{equation}
    \mathcal{L}_{CG} = -\frac{1}{T}\sum_{t=1}^T \log p(y^*_t|y^*_{<t},\hat{V},l,W^{\text{info}};\Theta),
\end{equation}
where $y^*_t$ denotes the $t$-th word of ground-truth caption for the current textbox location, and $\Theta$ denotes all learnable parameters of the pre-training model.
During the inference phase, we first encode the image, location and product information embeddings, and then feed a special start token \texttt{[SOS]} to predict the caption word by word.

\noindent\textbf{Task \#2: Caption Matching (CM).}
To help the model learn which captions are appropriate for the current image and location while which are not, we further introduce another pre-training task called Caption Matching.
It is similar to the ITM task commonly used in vision-and-language pre-training models \cite{chen2019uniter,li2020unicoder,lu2019vilbert,zhuge2021kaleido}, which requires the model to predict if the image and caption are semantically aligned.
A score $s$ between 0 and 1 is predicted by the hidden output of the \texttt{[SOS]} token.
The positive examples of this task are corresponding pairs in the dataset, while the negative examples can be diverse.
In this work, we design three levels of negative example construction and progressively learn the task from easy to difficult.

\noindent\textbf{Level-\uppercase\expandafter{\romannumeral1}: Image caption matching.}
The first negative level is to randomly replace the correct caption with descriptions of other images.
Therefore, it is not consistent with the current image content.
We expect the model can recognize such negative examples according to the visual image and product information, which are the easiest negative cases.
\begin{table*}[ht]
    \centering
    \caption{We report BLEU (B), METEOR (M), CIDEr and Diversity (D) scores for the captioning on image task on the CapOnImage2M dataset. Since the CapOnImage task is a newly proposed task in this work, we adapt conventional state-of-the-art image captioning models to this task by introducing text location and textual knowledge for comparison.} 
    \label{tab:sota_comparison}
    \small
    \begin{tabular}{l | c c c c c c| c c c c c c}
    \toprule
     & \multicolumn{6}{|c}{Validation} & \multicolumn{6}{|c}{Test} \\
    \cmidrule{2-13}
    Methods & B@1 & B@4 & M & CIDEr & D@1 & D@2 & B@1 & B@4 & M & CIDEr & D@1 & D@2  \\ 
    \midrule
    Up-down \cite{anderson2018butd}  & 15.51 & 9.79 & 7.93 & 108.11 & 48.73 & 41.22 & 13.48 & 8.71 & 7.89 & 99.52 & 46.53 & 39.51 \\
    Up-down \cite{anderson2018butd} w/ TextAttn & 20.49 & 13.54 & 11.52 & 181.04 & 65.24 & 56.27 & 18.94 & 12.30 & 10.81 & 166.26 & 65.72 & 56.48 \\
    M2 \cite{cornia2020meshed} w/ TextAttn & 34.18 & 24.31 & 18.14 & 273.35 & 63.29 & 53.43 & 31.63 & 22.11 & 17.71 & 265.22 & 62.81 & 54.19 \\
    RSTNet \cite{zhang21rstnet} w/ TextAttn & 33.42 & 23.91 & 17.28 & 267.60 & 64.12 & 54.53 & 30.54 & 20.94 & 17.20 & 259.29 & 63.10 & 53.84 \\
    M4C-Captioner \cite{sidorov20textcap} w/o copying & 36.46 & 27.08 & 20.19 & 296.73 & 65.69 & 55.69 & 35.73 & 26.24 & 19.8 & 287.61 & 64.31 & 55.01 \\
    M4C-Captioner \cite{sidorov20textcap} w/ copying & 35.98 & 28.35 & 20.58 & 299.31 & 65.98 & 55.03 & 36.23 & 27.15 & 20.01 & 288.35 & 64.03 & 55.23 \\
    \midrule
    baseline & 36.46 & 27.08 & 20.19 & 296.73 & 65.69 & 55.69 & 35.73 & 26.24 & 19.78 & 287.61 & 64.31 & 55.01 \\
    no-locations (baseline w/o locations) & 17.78 & 9.36 & 10.03 & 99.08 & 22.94 & 17.24 & 17.29 & 11.04 & 9.73 & 95.52 & 23.05 & 17.28 \\ 
    no-info (baseline w/o info) & 20.81 & 13.88 & 11.53 & 133.26 & 55.87 & 47.32 & 19.94 & 13.19 & 10.10 & 125.51 & 53.84 & 46.31 \\
    no-image (baseline w/o image) & 22.25 & 16.32 & 13.51 & 155.59 & 58.71 & 49.85 & 21.83 & 15.33 & 11.82 & 147.32 & 58.42 & 48.61 \\
    context (baseline + location context) & 37.69 & 27.98 & 21.91 & 313.64 & 70.25 & 60.54 & 37.02 & 27.23 & 21.14 & 305.50 & 70.98 & 60.90 \\
    full (context + multi-task) & \bf 41.77 & \bf 32.20 & \bf 24.52 & \bf 357.03 & 74.05 & 63.20 & \bf 40.95 & \bf 31.45 & \bf 23.49 & \bf 345.41 & 74.87 & 63.70 \\
    \midrule
    human & - & - & - & - & \bf 90.13 & \bf 75.53 & - & - & - & - & \bf 89.91 & \bf 74.18  \\
    \bottomrule
    \end{tabular}
\end{table*}

\noindent\textbf{Level-\uppercase\expandafter{\romannumeral2}: Location caption matching.}
The second negative level is to replace the caption with those in other locations of the same image.
It is more difficult than the Level-\uppercase\expandafter{\romannumeral1} because the negative caption exactly describes the current image but is not suitable for the current location.
For example, the product efficacy descriptions may be inappropriate to appear on the left corner of the product image, while the product brand is more suitable.
We expect the model can learn the relationship of texts and textbox locations according to the surrounding visual context.

\noindent\textbf{Level-\uppercase\expandafter{\romannumeral3}: Neighbor-location caption matching.}
Since the captions in neighbor locations are the most confusing samples, we further introduce the third negative level, where we randomly replace the caption with those in neighbor locations, including the nearest previous location and the nearest next location defined in Section~\ref{sec:neighbor location}.
It can be seen as a special case of Level-\uppercase\expandafter{\romannumeral2}, which limits the negative location to the neighboring locations and makes it more difficult to distinguish.

\noindent\textbf{Progressive training strategy.}
Since the three levels of CM task are from easy to difficult, inspired by the human learning procedure, we propose a progressive training strategy to dynamically adjust the proportion of each level.
Specifically, we randomly replace captions with 60\% probability to form negative samples and leave 40\% unchanged as positive ones.
The negative captions come from the three levels with $p_1$, $p_2$ and $p_3$ probabilities respectively, where $p_1+p_2+p_3=1$.
We vary the probabilities over the course of training, according to the following formula:
\begin{eqnarray}
    && p_1 = \text{min}(1, 2 \cdot step\_num^{-0.2}), \\
    && p_3 = \text{min}(1, step\_num \cdot 5000^{-1.5}), \\
    && p_2 = \text{max}(0, 1 - p_1 - p_3).
\end{eqnarray}
It corresponds to rapidly decreasing the probability of Level-\uppercase\expandafter{\romannumeral1} from 1 at the beginning and then slowly decreasing to 0, while  linearly increasing the probability of Level-\uppercase\expandafter{\romannumeral3} from a very small value to 1.
As a result, the probability of Level-\uppercase\expandafter{\romannumeral2} will increase first, and then decrease.
Overall, the training objective of the CM task can be expressed as follows:
\begin{equation}
    \mathcal{L}_{CM} = - \mathbb{E}_{(\hat{V},l,W^{\text{info}},Y)\sim\mathcal{D}}[r \log s + (1-r) \log (1-s)],
\end{equation}
where $s$ refers to the predicted matching score of a training sample and $r\in[0,1]$ is the ground-truth label indicating whether it is a negative or positive sample.

\section{Experiments}

We carry out experiments to evaluate the ability of models for captioning on image given a provided text layout on the CapOnImage2M dataset.
We evaluate the caption generation qualities from multiple aspects, including the \emph{accuracy} measurement against the references, and the \emph{diversity} measurement within an image.
Since the correct caption for each textbox location is not unique, we further evaluate the \emph{fitness} of generated captions to the corresponding textbox locations with respect to the caption length and type.
Besides, we also conduct human evaluations.

\subsection{Experimental Setup}

\noindent\textbf{Evaluation metrics.}
For the accuracy measurement, we evaluate the generated captions against the ground-truth with standard metrics used in the image captioning task, including BLEU \cite{papineni2002bleu}, METEOR \cite{denkowski2014meteor} and CIDEr \cite{vedantam2015cider}.
For the diversity measurement, we concatenate the dense captions within an image as a paragraph and compute the ratio of unique $n$-grams, called Div@$n$ \cite{shetty2017div1}.
For the fitness measurement, we show the relationship of generated caption length to the aspect ratio of text box, and the type distribution of generated captions on the image.

\noindent\textbf{Implementation details.}
We initialize the ResNet-50 backbone with the checkpoint pre-trained on ImageNet, and fine-tune it with our model in an end-to-end manner.
Our model has $L=6$ transformer layers with the hidden dimension of $d=1024$ and attention head $A=8$.
In the pre-training stage, we sample the batch of CG and CM tasks with a proportion of 3:1 for 200K steps.
We adopt a warming-up strategy for the first 4K steps.
For text processing, we tokenize Chinese captions into characters and build a vocabulary with 6263 tokens.

\subsection{Comparison with Baseline Models}
Since the CapOnImage task is a newly proposed task in this work, we adapt conventional state-of-the-art image captioning models (Up-down \cite{anderson2018butd}, RSTNet \cite{zhang21rstnet}, M2~\cite{cornia2020meshed}, M4C-Captioner \cite{sidorov20textcap}) to this task as the baselines for comparison.

\noindent\textbf{Up-down w/ TextAttn.}
Up-down \cite{anderson2018butd} model is a LSTM-based image captioning model, which contains an attention layer for visual context computation and a language layer for caption generation.
We enhance it with textual attention for the product information and concatenate the textual context with the visual context for caption generation.
We initialize the LSTM hidden state with the corresponding location embedding.

\noindent\textbf{$M^2$ w/ TextAttn.}
$M^2$ is a transformer-based image captioning method, which uses additional memory slot to store object relationship. 
We add an additional transformer encoder for product information and concatenate it with the visual input for captioning.

\noindent\textbf{RSTNet w/ TextAttn.}
RSTNet \cite{zhang21rstnet} is the state-of-the-art image captioning model, which uses adaptive attention on the visual and language context.
We enrich the encoding output with location embedding and product information embedding, which is encoded by an additional transformer-based text encoder.

\noindent\textbf{M4C-Captioner.}
M4C-Captioner \cite{sidorov20textcap} is a classic text-aware image captioning model, which generates informative image captions according to both visual image and additional OCR tokens.
It encodes the two modalities with a common transformer encoder and uses a copying mechanism to generate the out of vocabulary OCR tokens.
We replace the OCR tokens with product information, and add an additional input of location embedding.

\begin{table*}[ht]
    \caption{Captioning results with different pre-training tasks and strategies. \emph{fixed} denotes training the multi-level CM task with a fixed proportion, while \emph{progressive} denotes varying the proportion according to the degree of difficulty and training steps.}
    \label{tab:ab_task}
    \centering
    \small
    \begin{tabular}{c |c c c| c c | c c c c| c c c c}
    \toprule
    \multirow{2}{*}[-0.5ex]{Row} & \multicolumn{3}{c|}{Pre-training tasks} & \multicolumn{2}{c|}{Pre-training strategy} & \multicolumn{4}{c|}{Validation} & \multicolumn{4}{c}{Test} \\ 
    \cmidrule{2-14}
    & Level-I & Level-II & Level-III & fixed & progressive & B@1 &  B@4 & M & CIDEr & B@1 & B@4 & M & CIDEr \\
    \midrule
    1 & & & & - & - & 37.69 & 27.98 & 21.91 & 313.64 & 37.02 & 27.23 & 21.14 & 305.50 \\
    \midrule
    2 & \checkmark & & & \checkmark & & 38.87 & 29.23 & 22.32 & 325.74 & 38.19 & 28.48 & 21.69 & 316.46 \\
    3 & \checkmark & \checkmark & & \checkmark & & 40.22 & 30.36 & 22.88 & 339.72 & 39.36 & 29.54 & 22.03 & 329.77 \\
    4 & \checkmark & \checkmark & \checkmark & \checkmark & & 40.55 & 30.71 & 23.34 & 347.30 & 39.40 & 29.57 & 22.51 & 335.49 \\
    \midrule
    5 & \checkmark & \checkmark & \checkmark & & \checkmark & \bf 41.77 & \bf 32.20 & \bf 24.52 & \bf 357.03 & \bf 40.95 & \bf 31.45 & \bf 23.49 & \bf 345.41 \\
    \bottomrule
    \end{tabular}
\end{table*}
\begin{table}[ht]
    \centering
    \caption{Captioning results on the test set with different contextual locations. The number in () means how many locations used as the context to enhance the current location embedding.}
    \label{tab:ab_context}
    \small
    \begin{tabular}{l | c c c c c}
    \toprule
    Methods &B@1 & B@4& CIDEr & D@1 & D@2 \\ 
    \midrule
    w/o location context (0) & 35.73 & 26.24 & 287.61 & 69.31 & 59.01 \\
    w/ two random locations (2) & 33.98 & 24.79 & 283.73 & 67.43 & 57.82 \\
    w/ top-1 nearest locations (2) & 37.02 & \bf 27.23 & 305.50 & \bf 70.98 & \bf 60.90  \\
    w/ top-2 nearest locations (4) & \bf 37.88 & 27.10 & \bf 307.61 & 70.12 & 60.34 \\
    \bottomrule
    \end{tabular}
\end{table}

\noindent\textbf{Variants of our model.}
We also compare with different variants of our model.
Since all the words to be generated are already in the vocabulary, the copy mechanism bring no significant improvement (
Table~\ref{tab:sota_comparison}), so we remove it and use M4C-Captioner w/o copying as our   \emph{baseline} model. 
The \emph{no-info} model adopts the same architecture as the \emph{baseline} model except that the product information input is removed.
Similarly, the \emph{no-image} and the \emph{no-locations} model share the same baseline model architecture but with the image and locations input removed respectively. 
Our \emph{context} model is the \emph{baseline} model enhanced with neighbor location contexts, which is still trained only with the CG task.
The \emph{full} model is our complete model with progressive pre-training by both CG and CM tasks on the same dataset.

Table~\ref{tab:sota_comparison} reports the captioning on image results of different models on the CapOnImage2M validation and test sets.
It is shown that the conventional image captioning model without any adaptation perform poorly
on the CapOnImage task.
This is because these models lack the textual context that can provide rich information for caption generation. Enhancing the Up-down, $M^2$, and RSTNet with textual attentions on the additional product information, the captioning results are significantly improved.
However, they are still inferior to the adapted text-aware image captioning model, which has a good ability of multi-modal fusion with the cross transformer encoder.
Therefore, it stands for a strong baseline for our model.
Compared with the \emph{baseline} model, our \emph{context} model enhances the text location embedding with neighbor location contexts, which brings significant improvements on both accuracy and diversity metrics.
It demonstrates the importance of location relationship modeling especially for reducing caption redundancy.
Although good results have been achieved, the model is only trained with caption generation objective against the ground-truth, which is not sufficient to help the model learn complex correspondence between texts and image locations.
Therefore, when pre-training the \emph{context} model with both CG and CM tasks in a progressive manner, our \emph{full} model achieves the state-of-the-art results.
Nevertheless, there is still a gap with the human annotations on the captioning diversity metrics.

To further explore the contribution of each input modalities, we also report the captioning results with some input removed(\emph{no-locations}, \emph{no-info}, and \emph{no-image}).
It shows that the location information is more important than the visual image and the textual information for the CapOnImage task.
However, 
these three models are severely inferior to the \emph{baseline} model with multi-modal input, which shows the necessity of multi-modal fusion for this new task.


\subsection{Ablative Analysis}

\begin{figure}
  \centering
  \includegraphics[width=\linewidth]{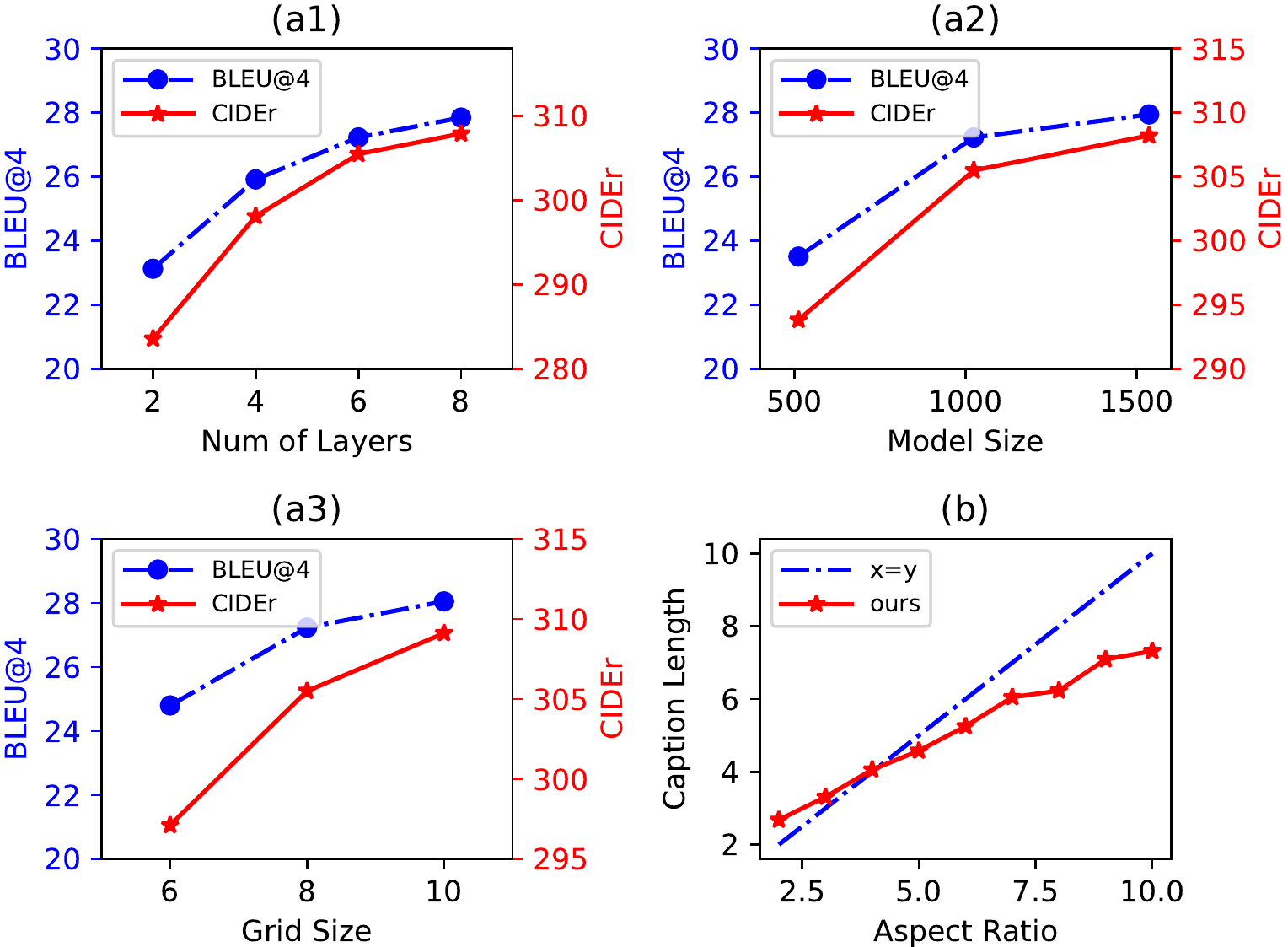}
  \caption{\emph{(a1)}: Ablation of number of transformer layer. \emph{(a2)}: Ablation of the hidden dimension of transformer model. \emph{(a3)}: Ablation of grid size of resnet backbone. \emph{(b)}:  The average captioning length of our model for the text box with different aspect ratios.}
  \label{fig:ablation}
  \vspace{-8pt}
\end{figure}

\noindent\textbf{Parameters determination.}
In 
Figure~\ref{fig:ablation}, we conduct ablation studies to investigate the suitable parameters for our model. We use our \emph{context} model and operate our experiment on test set. The parameter that need to be determined are number of layers of transformer,
hidden dimension of transformer and grid feature size of resnet backbone. In 
Figure~\ref{fig:ablation}(a), we study the impact of these three parameters for caption performance(BLEU@4 and CIDEr). We choose 6 layer transformer with hidden dimension 1024 and 8$\times$8 grid size resnet for the intuition of Accuracy-Efficiency Trade-Offs. 

\noindent\textbf{Pre-training tasks and strategy.}
In Table~\ref{tab:ab_task}, we ablate the proposed multi-level pre-training tasks and progressive training strategy.
It shows that pre-training with only Level-I of the CM task (row 2) can significantly improve the non-pretrained model with only CG objective (row 1).
It demonstrates the importance of multi-modal alignment to the CapOnImage task.
Upgrading the CM task with more difficult negatives in Level-II helps the model better learn the relationship of captions and text locations and thus yields better results (row 3). Further incorporating negatives in Level-III bring additional gains (row4), which confirms the importance of context information in CapOnImage task.
However, since three levels of negative samples are built from easy to difficult, we seek to boost the learning process of CM task in an adaptive fashion: the ratio of pre-training tasks need to be adapted to the training status and vary in a progressive manner(as opposed to a fixed proportion of 30\%:40\%:30\%).
Therefore, we propose a progressive training strategy with the proportion of easy task decreased and hard task increased in the training process.
It boosts the results stably (row 5).

\noindent\textbf{Choice of contextual locations.}
In Table~\ref{tab:ab_context}, we take a further study on the neighbor-enhanced location embedding module with different contextual locations.
With the nearest neighbor (previous and next) locations used as the context as described in Section~\ref{sec:neighbor location}, our model significantly improves the accuracy and diversity metrics.
To figure out where the benefit comes from, we compare with the model using the same amount of randomly selected locations as context.
Experimental results show that the randomly selected locations cannot improve the results and may even bring noise, which demonstrates the effectiveness of our model in encoding neighboring layout information to generate more appropriate and diverse captions.
When further expanding the contextual range from the top-1 nearest to the top-2 nearest (top-2 previous and top-2 next), the model achieves slightly better result on the accuracy metric.
To balance the efficiency and quality, we finally use the top-1 nearest locations as the context in our model.

\begin{figure*}
  \centering
  \includegraphics[width=0.91\linewidth]{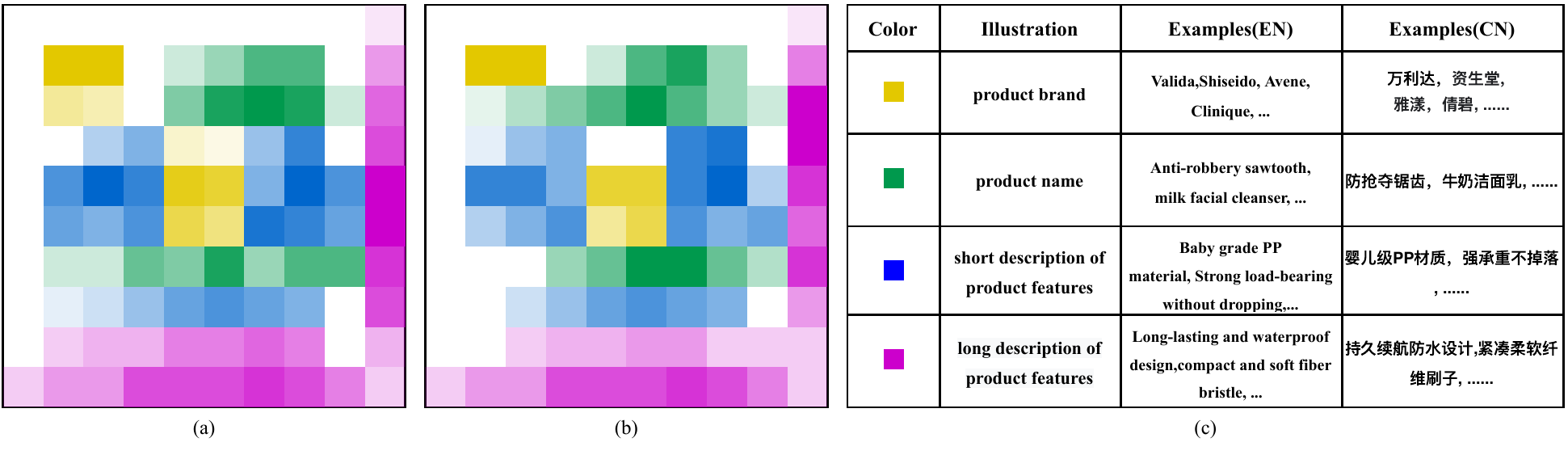}
  \vspace{-8pt}
  \caption{\emph{(a)}:Distribution of the GT caption types. \emph{(b)}: Distribution of generated caption types. \emph{(c)}:  Illustration of the four caption types in CapOnImage2M dataset via automatic clustering.}
  \label{fig:type}
\end{figure*}

\subsection{Text-Location Fitness}
\noindent\textbf{Caption length to textbox aspect ratio.}
Given a textbox location, the generated caption should exactly fit in with it for visual aesthetic.
The text length and font size are the influencing factors.
Since the short side of the textbox determines the font size, the aspect ratio (long side length~/~short side length) can reflect the most suitable text length.
Therefore, we show the relationship of our generated caption length with aspect ratio of the corresponding textbox in the Figure~\ref{fig:ablation}(b).
It shows that with the textbox aspect ratio increased, our model generates longer captions almost linearly, which demonstrates the controllability of the text box size to the length of the caption generated by our model.

\noindent\textbf{Caption type to textbox location.}
Besides the caption length, the types of captions in CapOnImage2M dataset are diverse at different image locations.
To explore whether the type of our generated captions is suitable to the given locations, we visualize the caption type distribution on the image.
Since the caption type annotations are not available, we automatically group the ground-truth captions into 4 categories by k-means based on their sentence-level BERT \cite{devlin2019bert} embeddings.
We then display the same type of captions using the same color on an image. 
As shown in Figure~\ref{fig:type}(a), the captions with the same type are located together, which shows that the caption type is very related to its location. We assign our generated captions to the 4 clusters and visualize them in the same way in Figure~\ref{fig:type}(b).
It looks very similar to the ground-truth type distribution map, which shows that our model effectively learns the relationship of text location and text type.
The meaning of each color are illustrated in Figure~\ref{fig:type}(c). 
\begin{table}[t]
    \caption{Human evaluation of our \emph{full} model vs. \emph{baseline} model on the test set w.r.t. relevance, diversity and informativeness.}
    \label{tab:human_eval}
    \centering
    \small
    \begin{tabular}{l c c c}
    \toprule
     & Base wins (\%) & Full wins (\%) & Delta \\
    \midrule
    relevance & 15.93 & 34.50 & \bf +18.57  \\
    diversity & 14.36 & 36.88 & \bf + 22.52  \\
    informativeness & 18.86 & 40.19 & \bf +21.33 \\
    \bottomrule
    \end{tabular}
\end{table}
\vspace{-8pt}
\subsection{Human Evaluation}
In addition to the objective evaluation, we also conduct human evaluation on 400 randomly sampled images from the test set.
We render the generated dense captions from \emph{baseline} model and our \emph{full} model on the image via opencv.
We instruct workers to choose which one is better or they are not distinguishable based on relevance, diversity and informativeness respectively.
Each image was judged by two different workers.
To avoid the prior bias, we anonymize the model names and shuffle the predictions randomly.
Table~\ref{tab:human_eval} shows the human evaluation results.
Our \emph{full} model significantly 
outperforms the \emph{baseline} model especially on the diversity and informativeness aspects, which demonstrates the effectiveness of the proposed neighbor-enhanced location embedding and multi-level progressive pre-training.

\subsection{Qualitative Results}
\begin{figure}
  \centering
  \includegraphics[width=\linewidth]{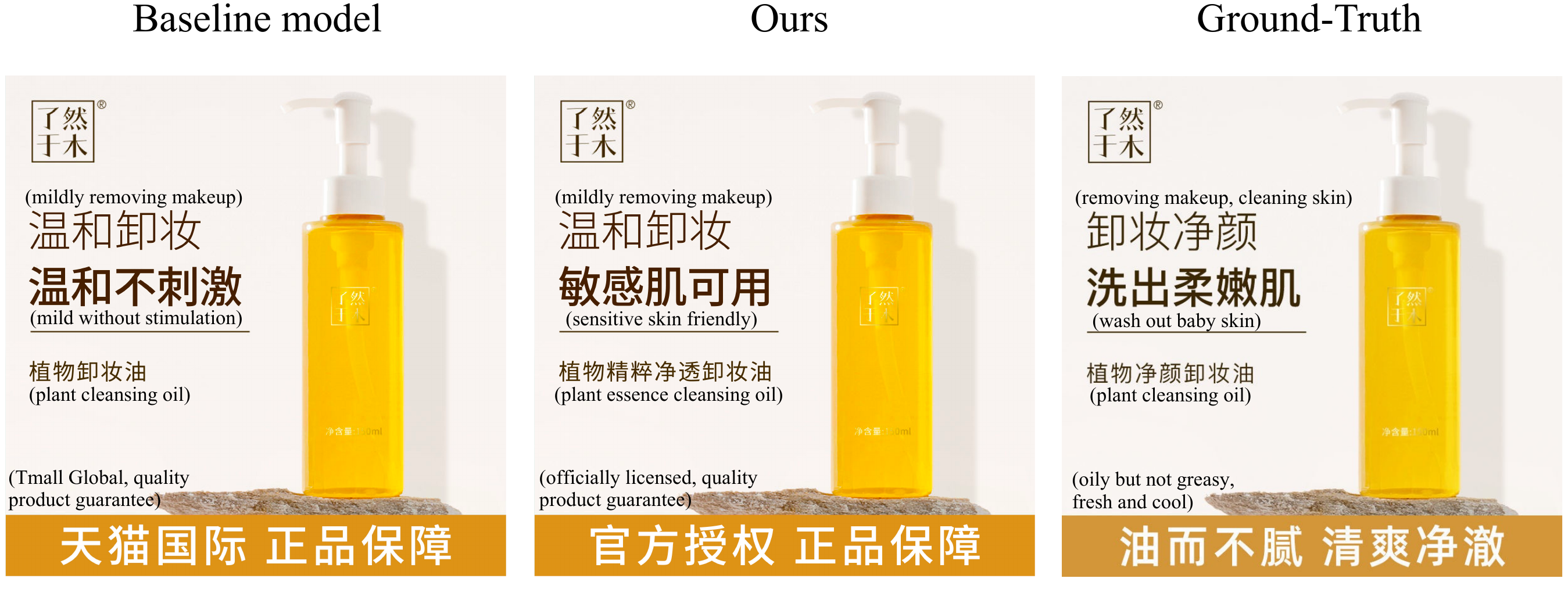}
  \includegraphics[width=\linewidth]{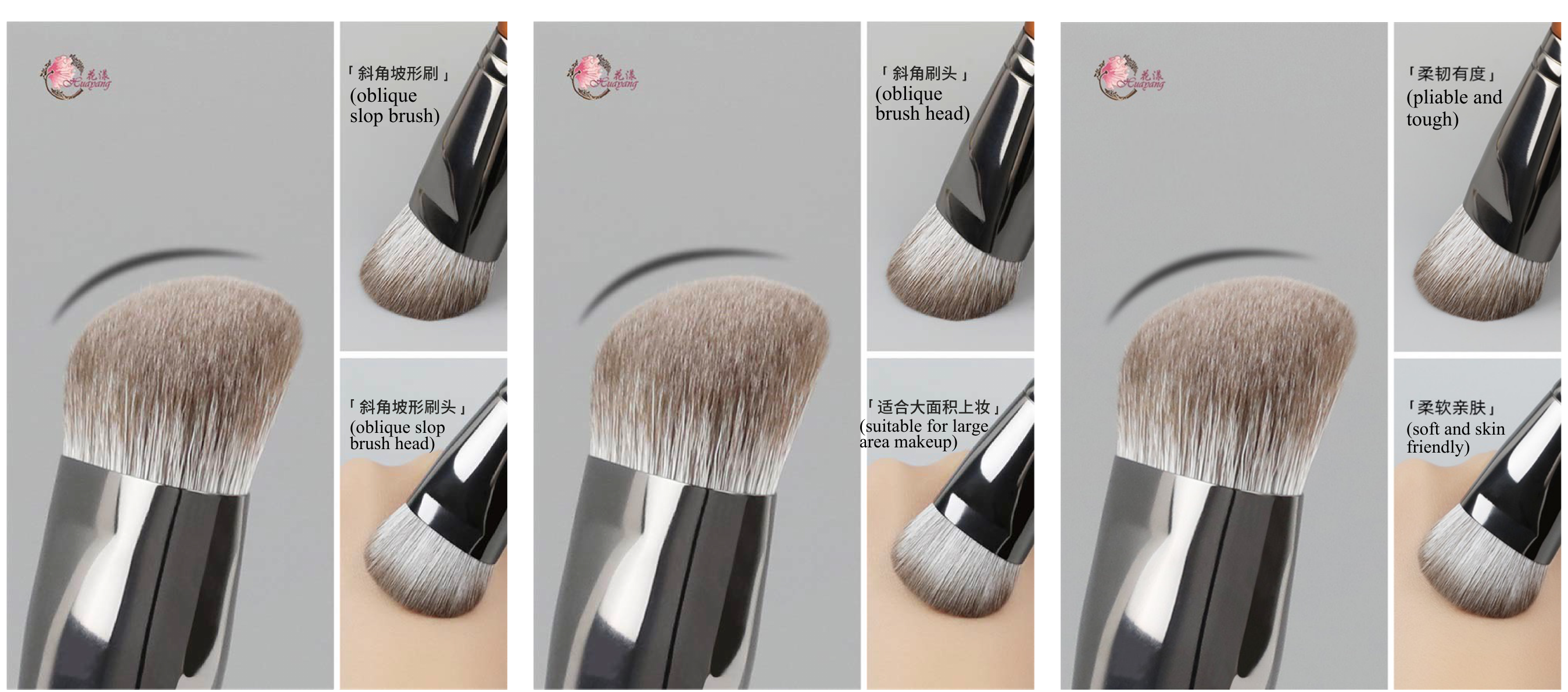}
  \caption{Qualitative dense-captioning on image results of our \emph{full} model and \emph{baseline} model. We add the English translation for each Chinese caption for better comprehension.}
  \label{fig:visualization}
  \vspace{-8pt}
\end{figure}
Figure~\ref{fig:visualization} visualizes some results of our \emph{full} model and \emph{baseline} model.
The \emph{baseline} model is shown to generate repetitive captions due to the lack of global layout awareness.
For example, for adjacent locations, the \emph{baseline} model repeats the concept of ``mild'', while our model generates more informative caption of ``sensitive skin friendly''.
Furthermore, our model is also shown to better exploit the visual context to generate more suitable captions.
In the second example, our model generates the text ``suitable for large area makeup'' for the down-right region where a hand appears, while the \emph{baseline} model fails to distinguish it with the up-right region and generates similar descriptions about the ``oblique slop brush''.
More visualization results can be found in the supplementary material.
\section{Conclusion}
In this work, we propose a new vision-and-language task called CapOnImage, which aims to generate dense captions at different locations on an image with visual and textual context.
We propose a multi-modal pre-training and fine-tuning model with multi-level pre-training tasks from easy to difficult for the correspondence learning between image location and text, and enhance the current location embedding with neighboring locations to reduce captioning redundancy.
Experimental results on the new proposed CapOnImage2M dataset show that our model achieves the state-of-the-art captioning results on both accuracy and diversity aspects compared with other image captioning model variants.
It is also shown that our model can generate controllable length and type of captions at different image locations.
In the future work, we will explore to generate dense captions with self-predicted text layout and combine the layout generation with caption generation in one joint framework to benefit from each other.

\bibliographystyle{ACM-Reference-Format}
\bibliography{sample-base}

\appendix
\section{Datasheet for CapOnImage2M Dataset}

\subsection{Motivation}

\begin{figure*}[ht]
    \centering
    \includegraphics[width=\linewidth]{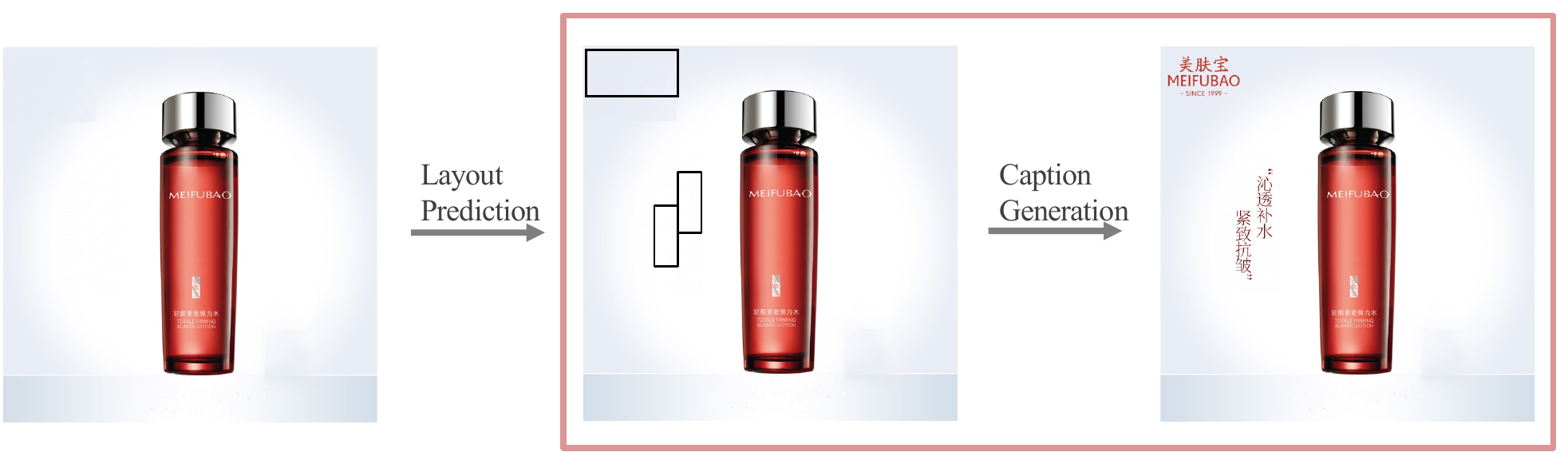}
    \caption{Pipeline of our CapOnImage task. We focus on generating captions for a provided list of text box locations in this work.}
    \label{fig:pipeline}
\end{figure*}

\noindent\textbf{For what purpose was the dataset created?}
The dataset was created to support the research on the captioning on image (CapOnImage) task, which aims to generate informative captions at different appropriate locations in the given image.  CapOnImage is a valuable task for both vision-and-language research and industrial applications. We show the pipeline of our task on Figure~\ref{fig:pipeline}.

\subsection{Composition}

\begin{figure}
    \centering
    \includegraphics[width=\linewidth]{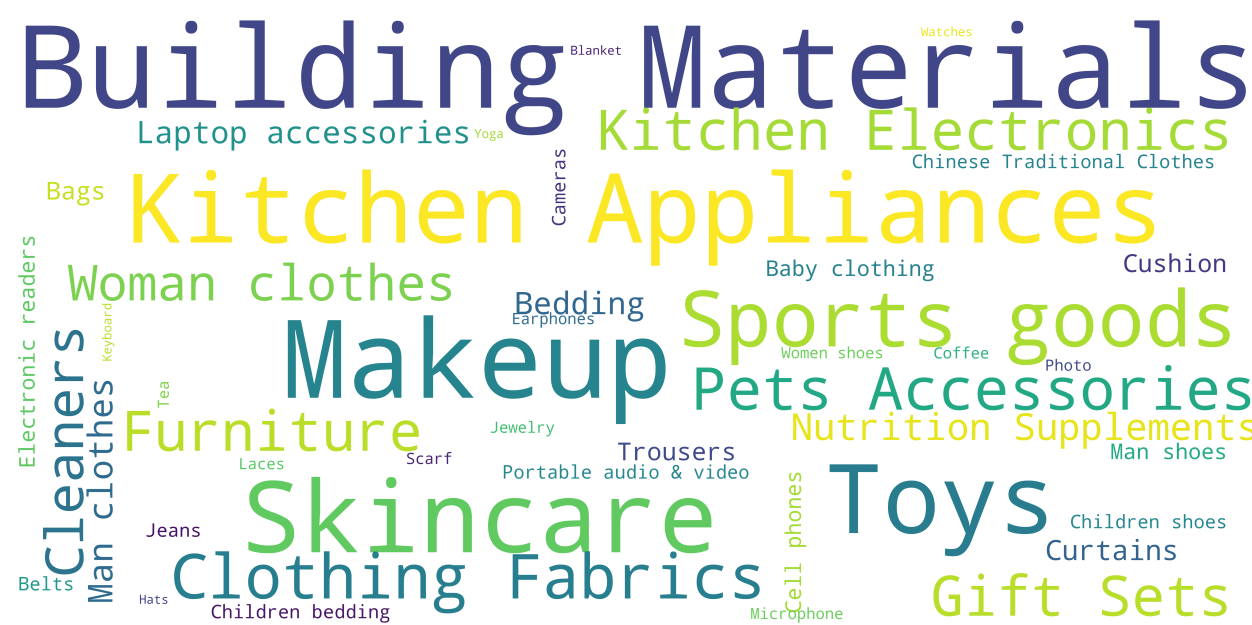}
    \caption{Word Cloud of 50 product categories. We show the Word Cloud according to the instances number of each category.}
    \label{fig:word_cloud_cate}
\end{figure}

\begin{figure*}
    \centering
    \includegraphics[width=\linewidth]{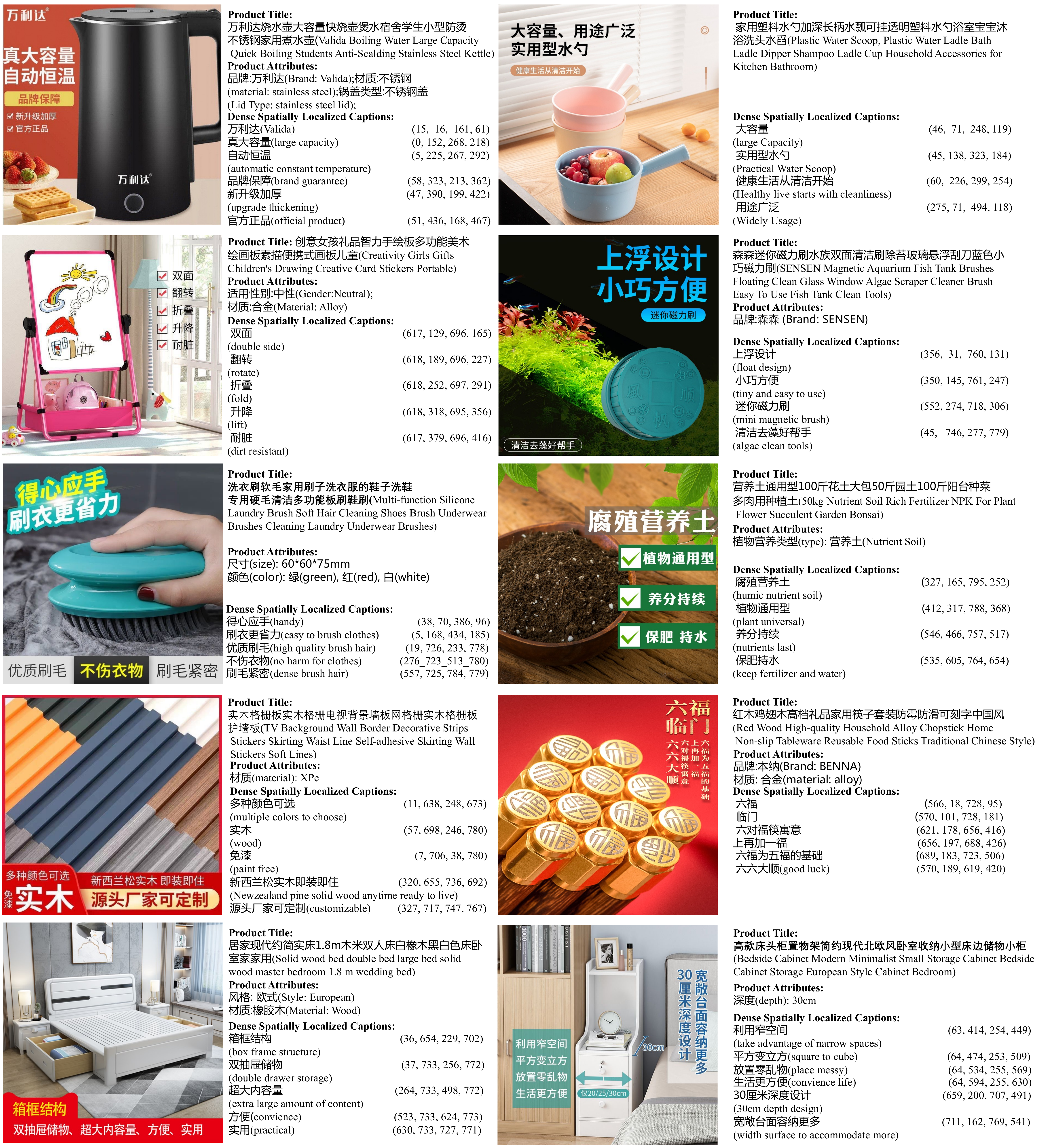}
    \caption{Examples of the CapOnImage2M dataset. We add the English translation for better understanding.}
    \label{fig:dataset_cases}
\end{figure*}

\noindent\textbf{What do the instances that comprise the dataset represent?}
Each instance in the CapOnImage2M dataset contains 50 categories product image, a sentence of product title, several product attributes, and multiple spatially localized captions with bounding box coordinates, describing the product from multiple aspects. We show the word cloud of product categories on Figure~\ref{fig:word_cloud_cate}.
Figure~\ref{fig:dataset_cases} shows some examples of the CapOnImage2M dataset.

\vspace{0.5em}
\noindent\textbf{How many instances are there in total?}
The dataset consists of 2.1M images and 10.07M texts in total. Each image contains an average of 4.8 spatially localized captions.

\vspace{0.5em}
\noindent\textbf{Does the dataset contain all possible instances or is it a sample (not necessarily random) of instances from a larger set?}
CapOnImage2M is a new independent dataset. The instances in CapOnImage2M dataset are crawled from an e-commercial website. New product images with texts will continue to emerge. Therefore, the current version of the dataset does not contain all possible instances.

\vspace{0.5em}
\noindent\textbf{What data does each instance consist of? Raw data (e.g., unprocessed text or images) or features?}
Each instance contains a product image with short side as 256 pixel in PNG format, the processed product title and captions.
The original image with high resolution can be downloaded by the provided image URL.

\vspace{0.5em}
\noindent\textbf{Is any information missing from individual instances?}
No. Everything is included in the dataset.

\vspace{0.5em}
\noindent\textbf{Are there recommended data splits (e.g., training, development/validation, testing)?}
Yes. The dataset is split into 2.06M images for training, 20K for validation, and 20K for testing.

\vspace{0.5em}
\noindent\textbf{Are there any errors, sources of noise, or redundancies in the dataset?}
We have removed redundant or similar images whose captions are overlapped over a threshold. Therefore, the instance redundancy in the dataset is low.
The captions are automatically recognized by an OCR model, therefore, there are noise in the captions of training set.
To ensure the evaluation accuracy, we further manually clean the OCR errors for the validation and testing sets.

\vspace{0.5em}
\noindent\textbf{Is the dataset self-contained, or does it link to or otherwise rely on external resources (e.g., websites, tweets, other datasets)?}
The dataset is self-contained, except that the original high resolution images are linked to a public website.
Nevertheless, the self-contained images are enough to reproduce the results, and the high resolution images are used for better visualization.

\vspace{0.5em}
\noindent\textbf{Does the dataset contain data that might be considered confidential (e.g., data that is protected by legal privilege or by doctorpatient confidentiality, data that includes the content of individuals non-public communications)?}
No. All data was collected from a publicly available e-commercial website.

\vspace{0.5em}
\noindent\textbf{Does the dataset contain data that, if viewed directly, might be offensive, insulting, threatening, or might otherwise cause anxiety?}
No. The dataset only consists of product images.

\vspace{0.5em}
\noindent\textbf{Does the dataset relate to people?}
Most of the images only contain products, and only a few images contain public product spokesmen.

\vspace{0.5em}
\noindent\textbf{Does the dataset contain data that might be considered sensitive in any way (e.g., data that reveals racial or ethnic origins, sexual orientations, religious beliefs, political opinions or union memberships, or locations; financial or health data; biometric or genetic data; forms of government identification, such as social security numbers; criminal history)?}
No. The dataset does not contain confidential information since all information was crawled from a public e-commercial website.

\begin{figure*}[ht]
  \begin{minipage}[ht]{0.5\linewidth}
    \centering
    \includegraphics[scale=0.27]{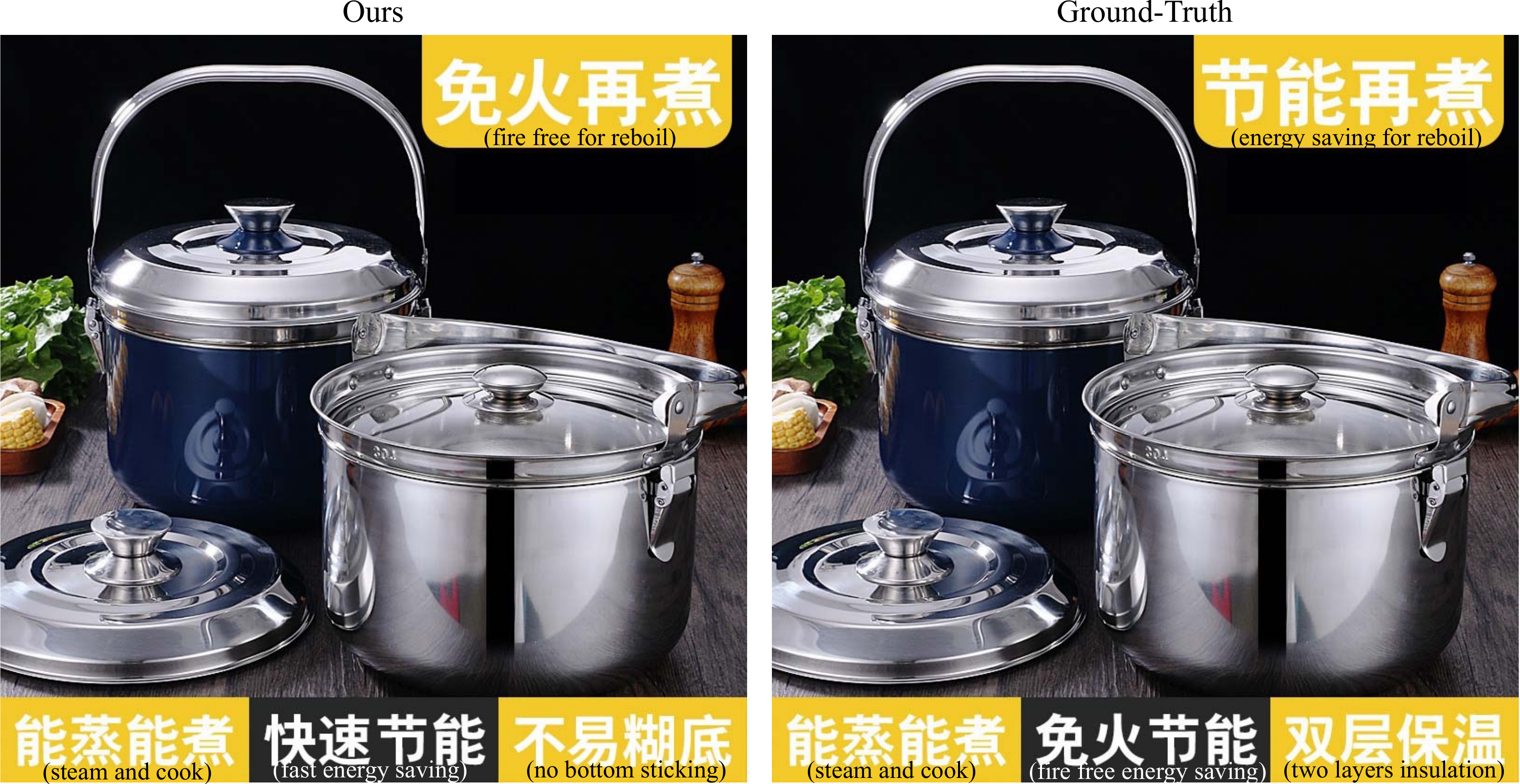}
  \end{minipage}%
  \begin{minipage}[ht]{0.5\linewidth}
    \centering
    \includegraphics[scale=0.27]{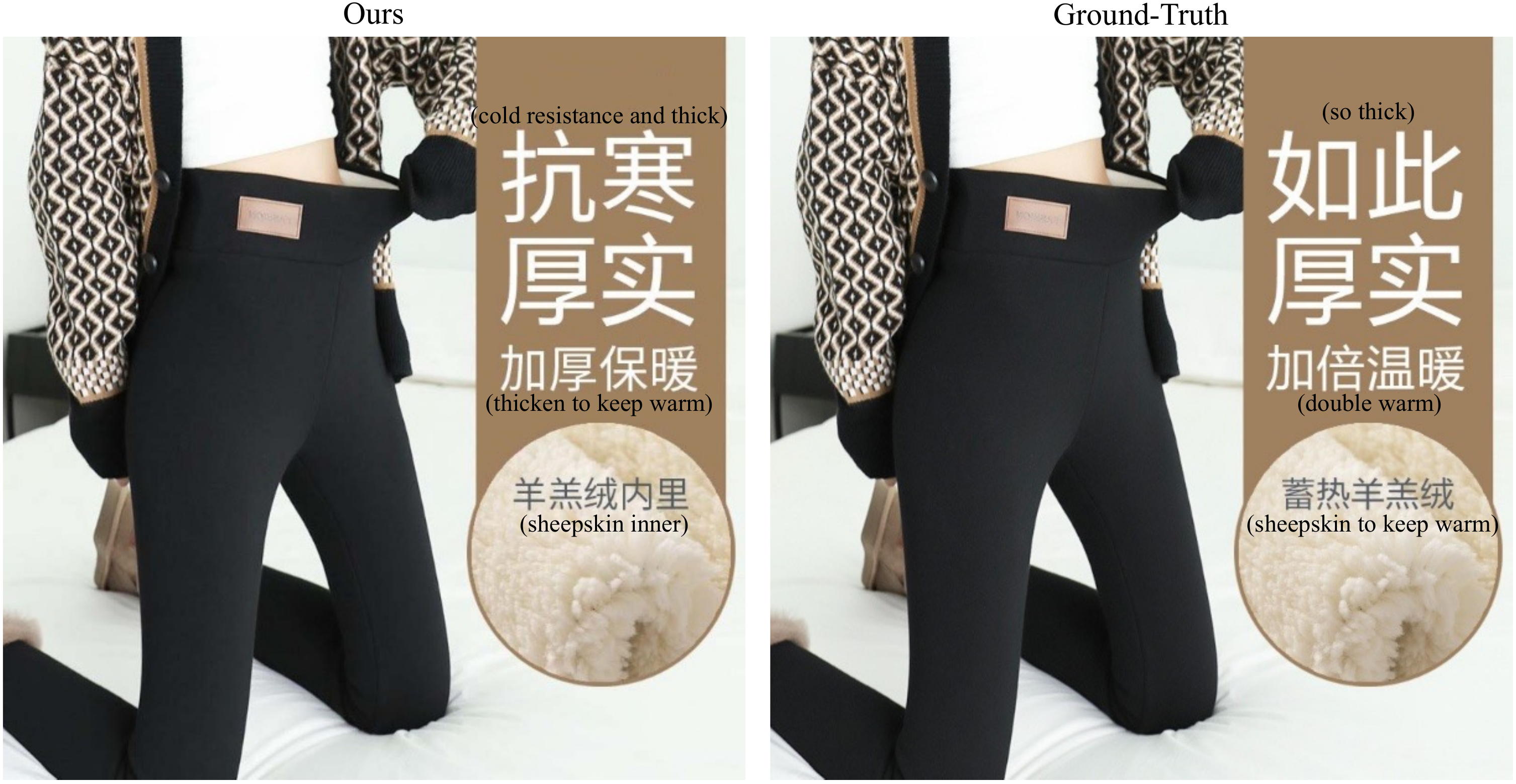}
  \end{minipage}
  \begin{minipage}[ht]{0.5\linewidth}
    \centering
    \includegraphics[scale=0.27]{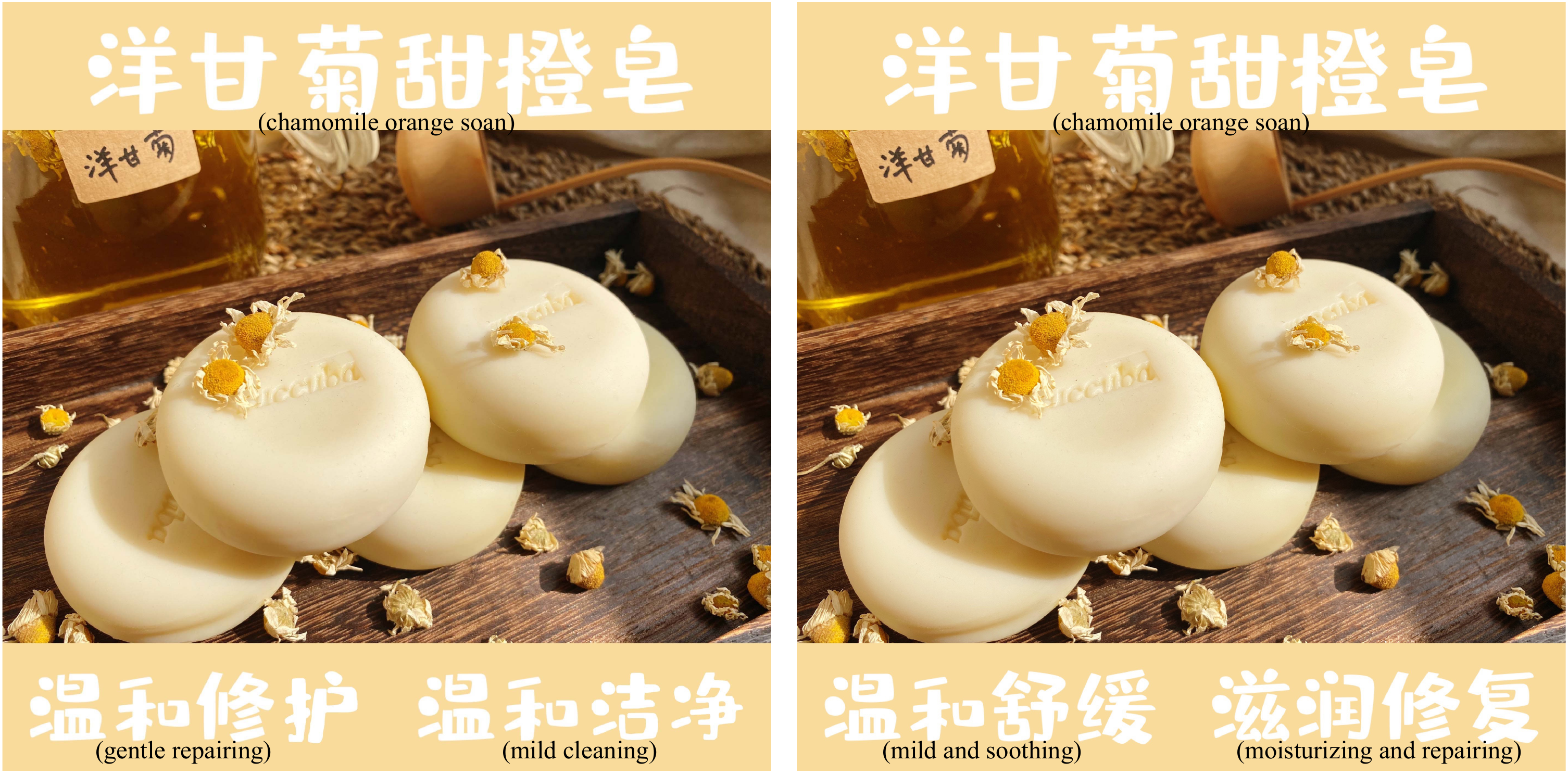}
  \end{minipage}%
  \begin{minipage}[ht]{0.50\linewidth}
    \centering
    \includegraphics[scale=0.27]{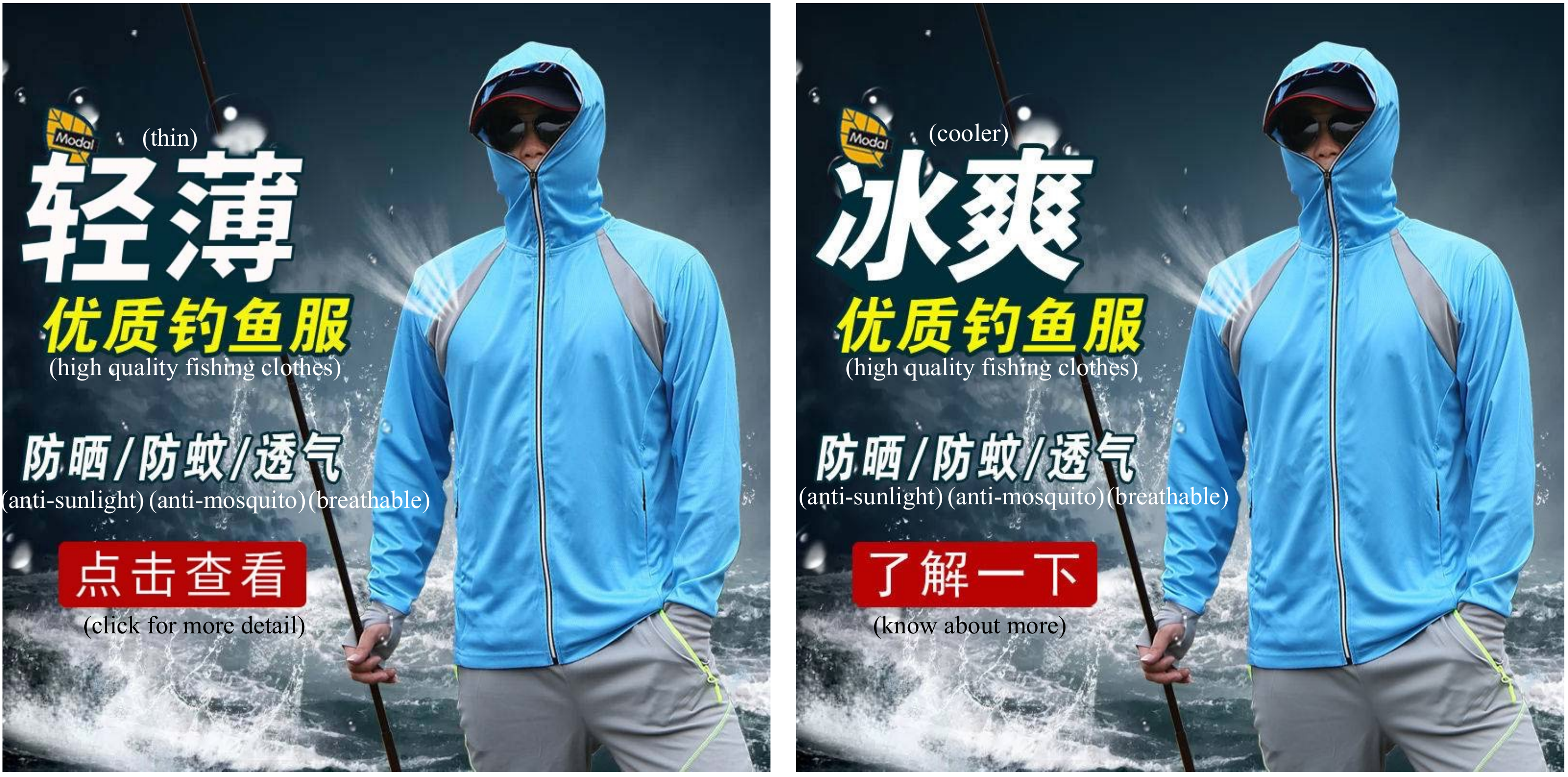}
  \end{minipage}
  \begin{minipage}[ht]{0.50\linewidth}
    \centering
    \includegraphics[scale=0.27]{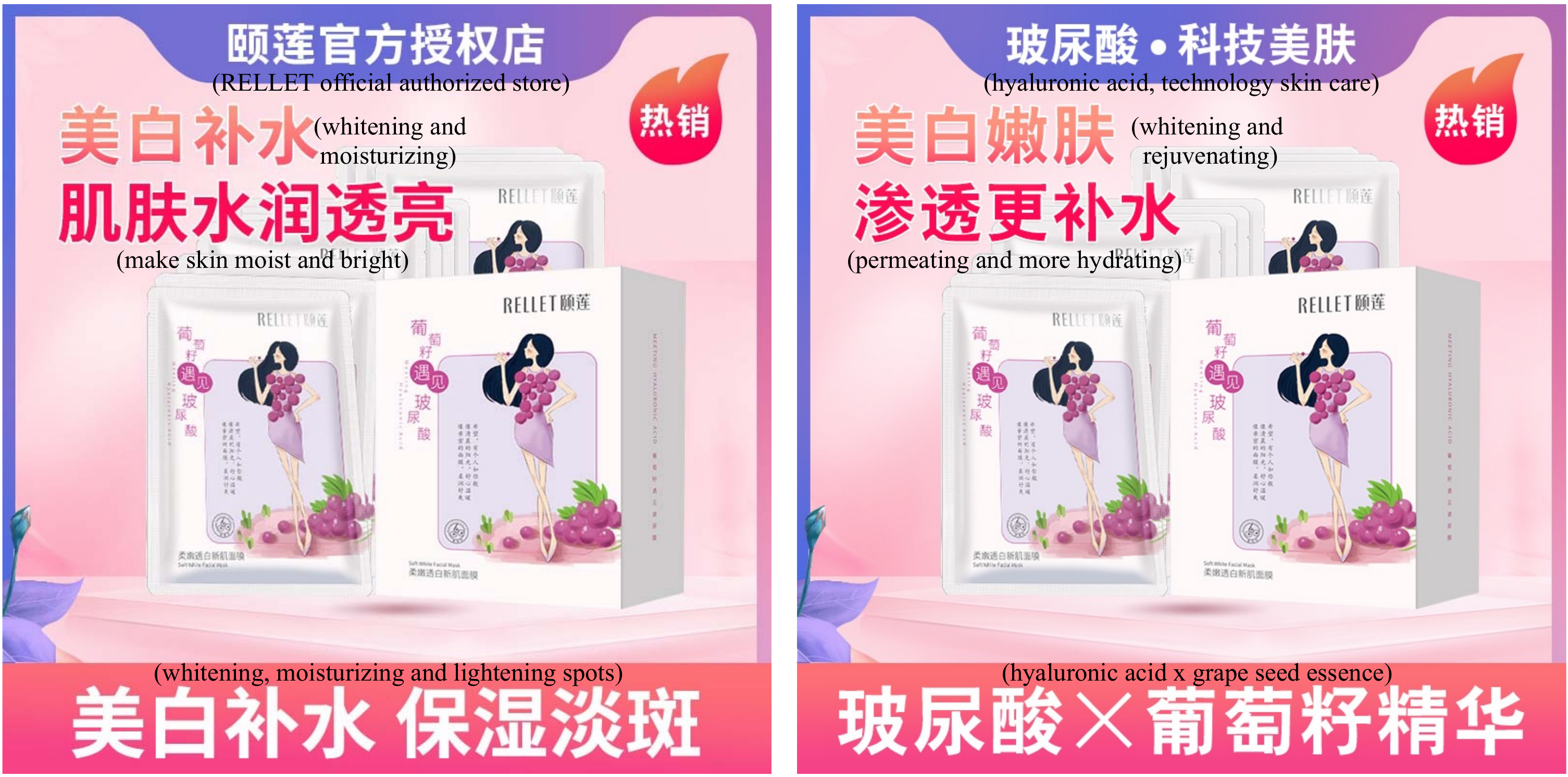}
  \end{minipage}%
  \begin{minipage}[ht]{0.50\linewidth}
    \centering
    \includegraphics[scale=0.27]{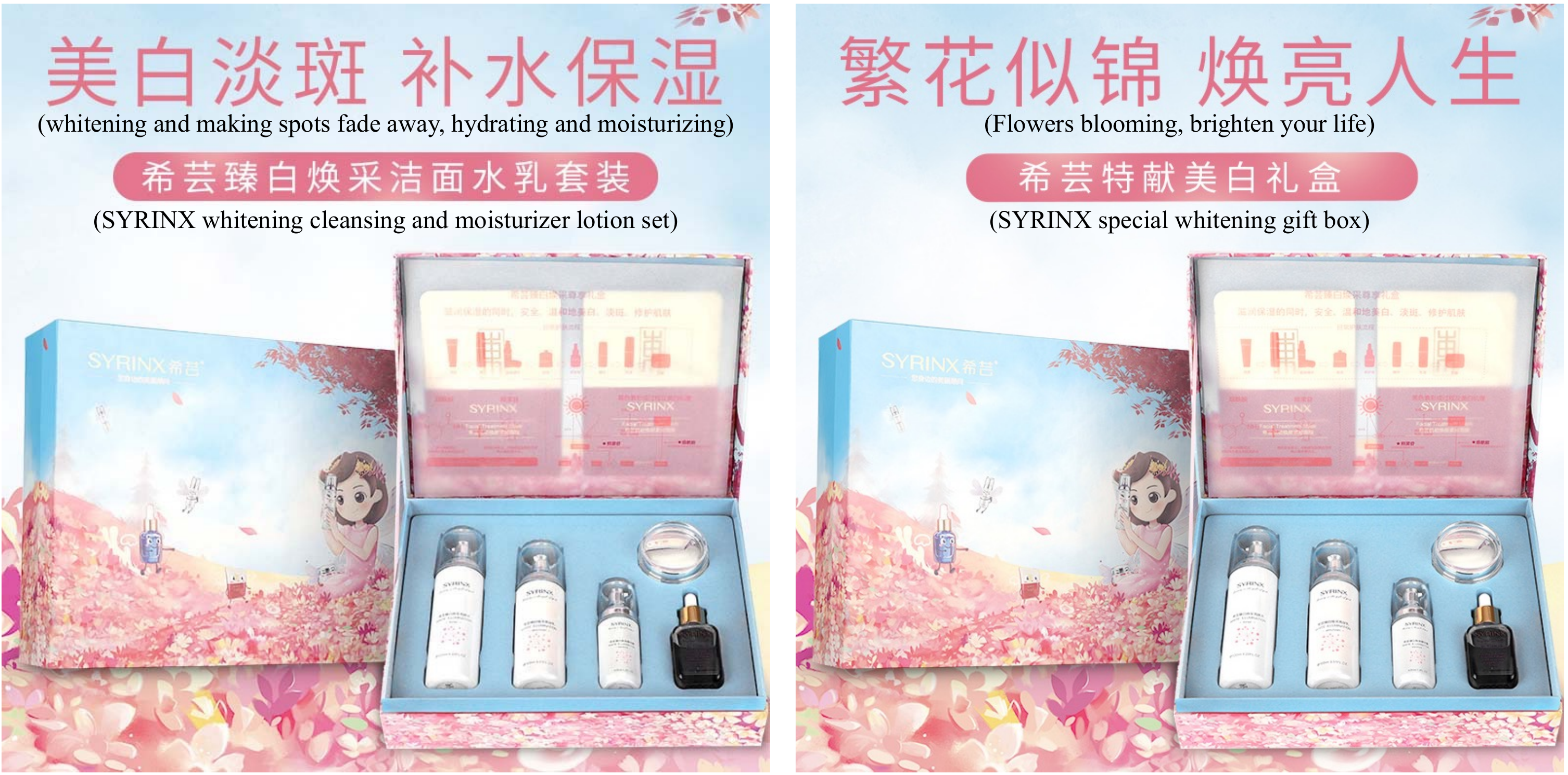}
  \end{minipage}
  \begin{minipage}[ht]{0.50\linewidth}
    \centering
    \includegraphics[scale=0.27]{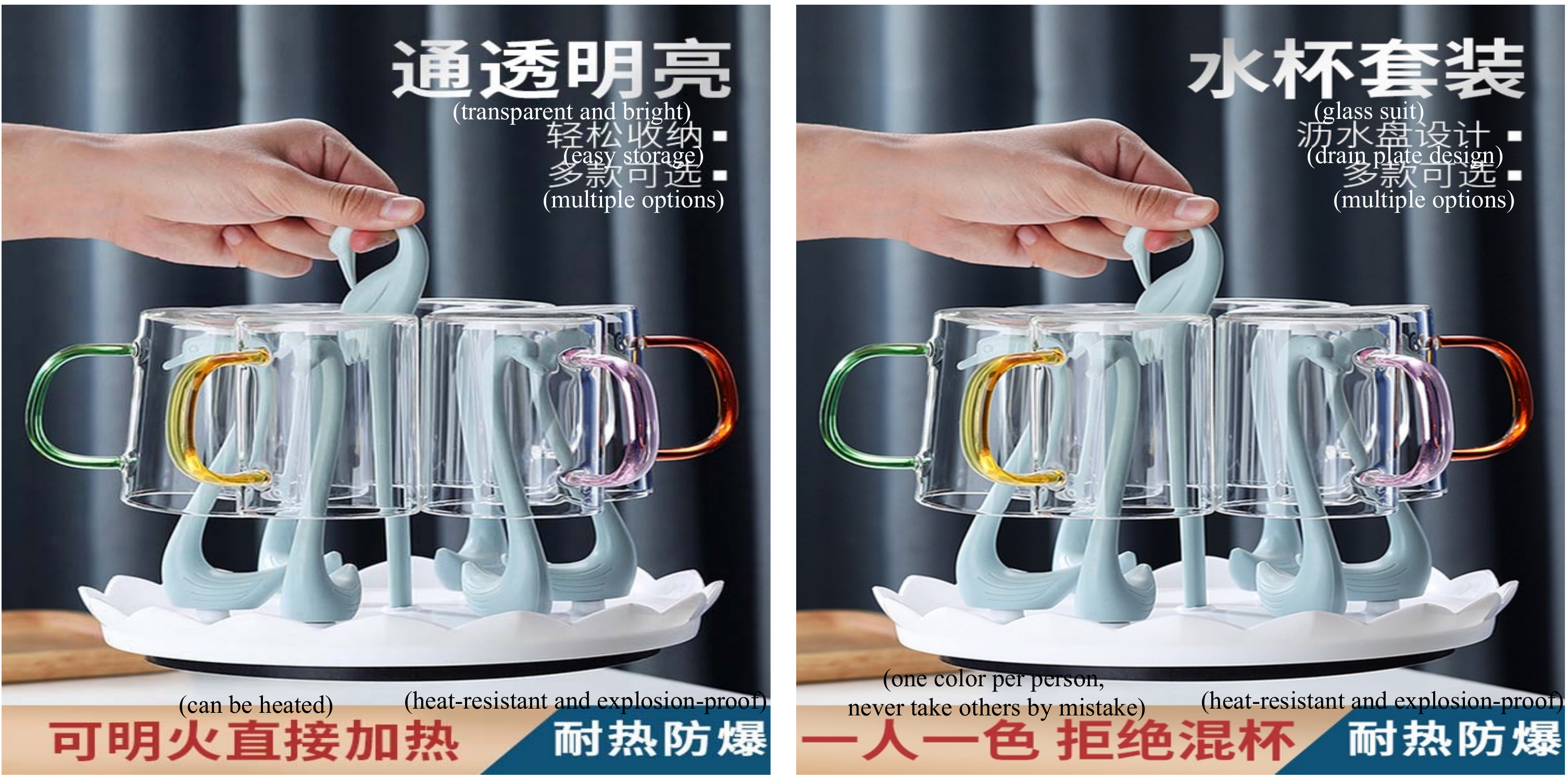}
  \end{minipage}%
  \begin{minipage}[ht]{0.50\linewidth}
    \centering
    \includegraphics[scale=0.27]{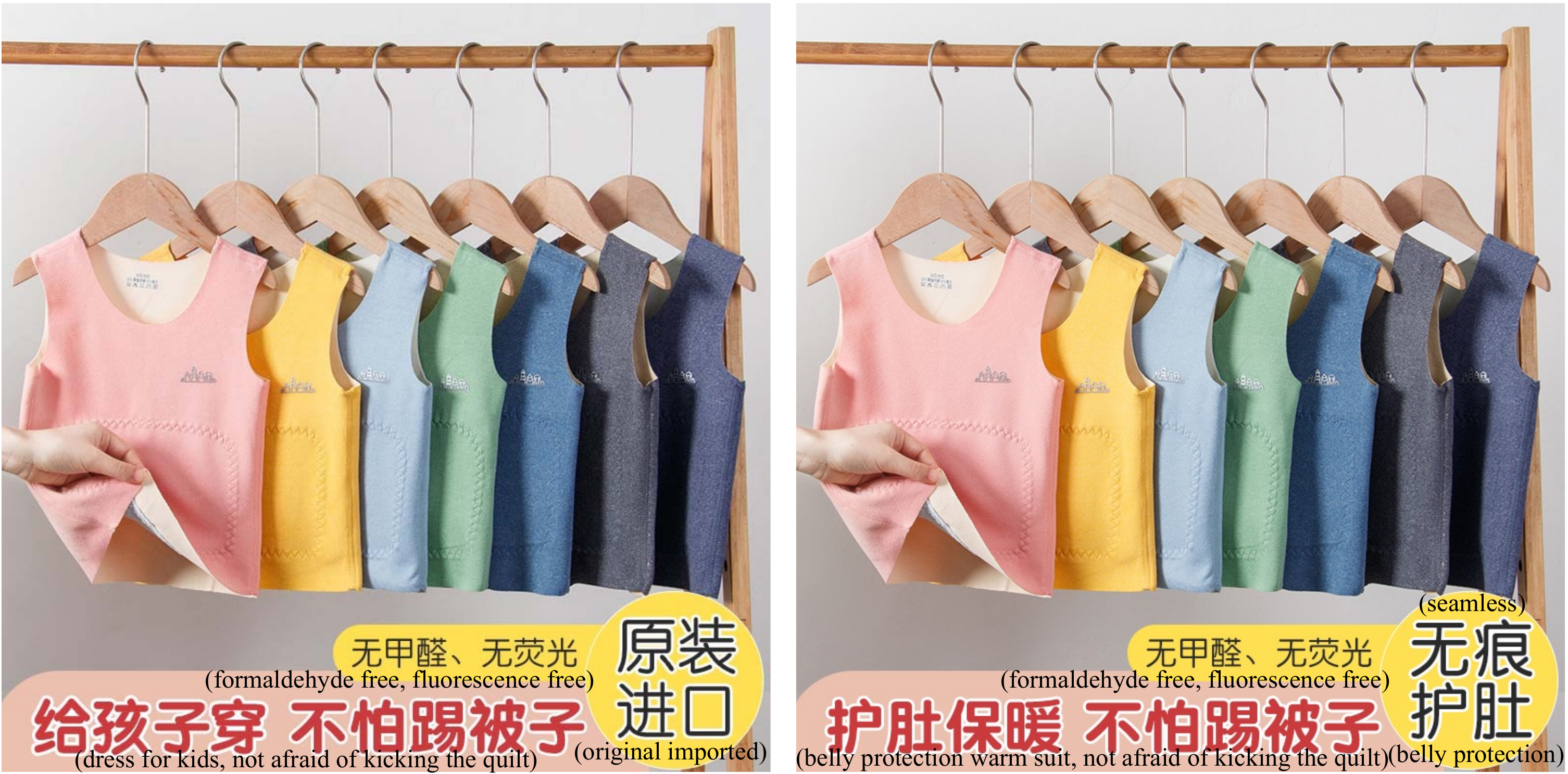}
  \end{minipage}
  \begin{minipage}[ht]{0.50\linewidth}
    \centering
    \includegraphics[scale=0.27]{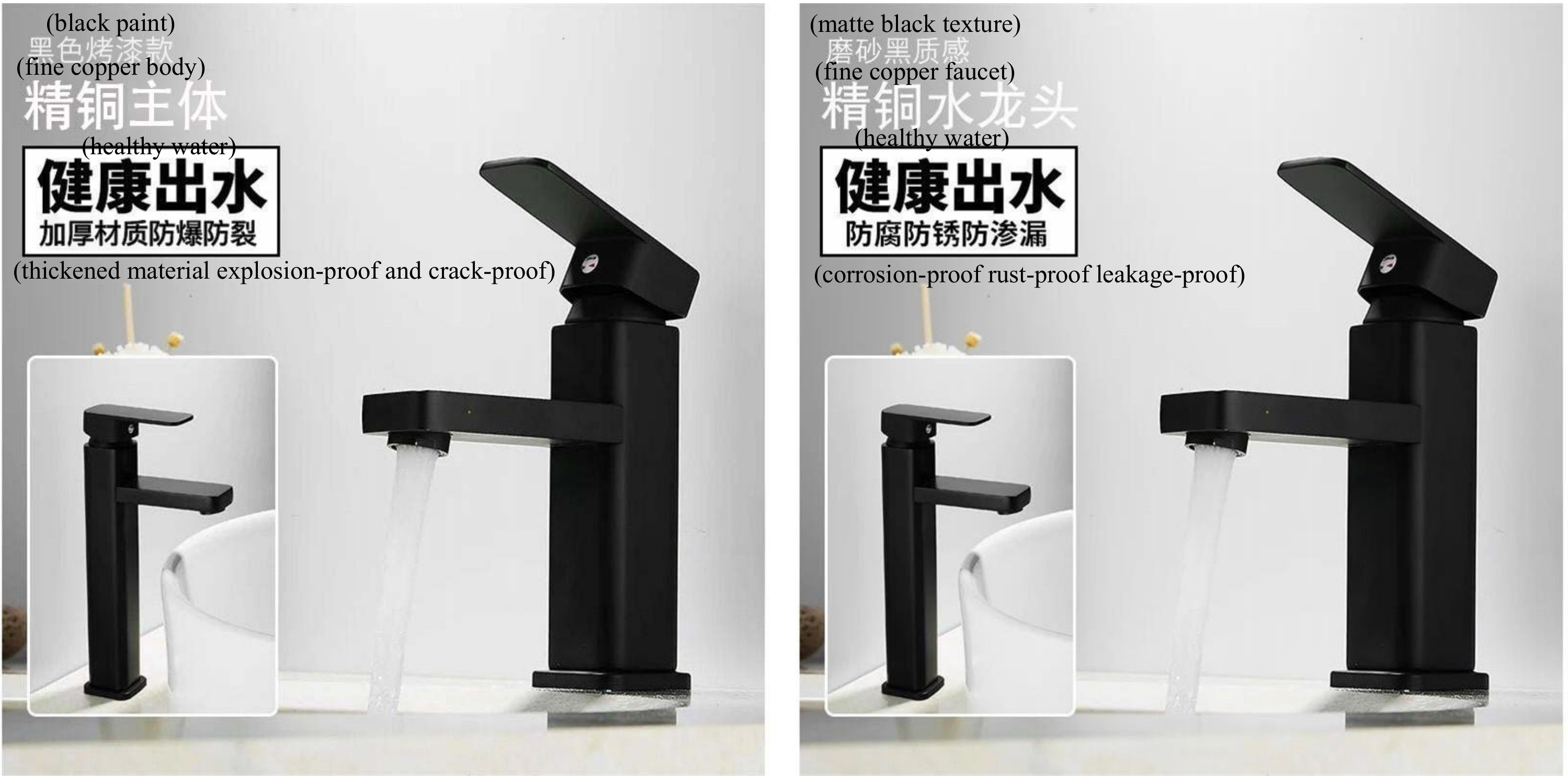}
  \end{minipage}%
  \begin{minipage}[ht]{0.50\linewidth}
    \centering
    \includegraphics[scale=0.27]{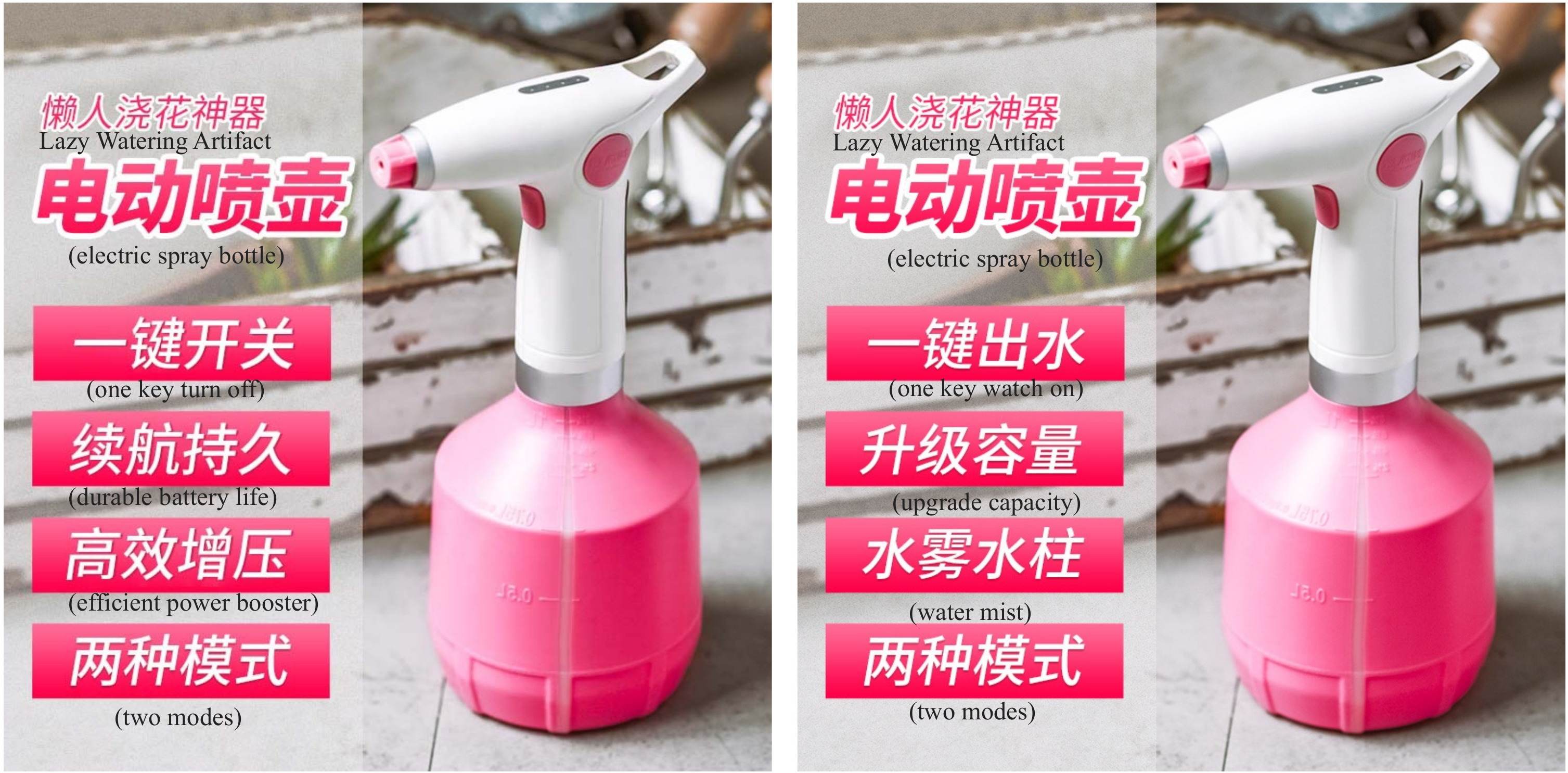}
  \end{minipage}
 
  \caption{Qualitative results of our \emph{full} model on the CapOnImage2M test set. We add the English translation for better understanding.}
  \label{fig:demo1}
\end{figure*}

\subsection{Collection Process}

\noindent\textbf{What mechanisms or procedures were used to collect the data?}
The raw images and product title sentences in this dataset were automatically crawled from Taobao website \footnote{https://taobao.com}.
The spatially localized captions are extracted from the image by an OCR model, and further manually cleaned for the validation and testing sets.

\subsection{Preprocessing}

\noindent\textbf{Was any preprocessing/cleaning/labeling of the data done?}
The following steps were taken to process the data: (1) \textit{Crawling raw images and product titles}. We first crawl the raw product images and titles from the e-commercial website, and then resize the images with short side as 256.
(2) \textit{Detecting texts on the image}. For each image, we employ an OCR toolkit to automatically detect the texts on the image as well as their bounding box coordinates.
(3) {\textit{Removing redundant instances}}. We remove redundant instances whose images contain similar captions that exceed the overlap threshold.
(4) {\textit{Removing discount information}}. We remove redundant instances that have high correlation with discount information.
(5) {\textit{Cleaning instances}}. We remove the texts longer than 10 or shorter than 2 characters, and remove the instances with only one caption on the image. Then, we input the automatically recognized captions into a pre-trained GPT model, and remove the captions with high generation perplexities.
(6) {\textit{Manual labeling}}. We further manually clean the captions with OCR errors in the validation and testing sets for accurate evaluation.

\vspace{0.5em}
\noindent\textbf{Is the software used to preprocess/clean/label the instances available?}
Yes. All software used to process the data is open source and has been mentioned above.

\subsection{Uses}

\noindent\textbf{Has the dataset been used for any tasks already?}
No, the dataset is newly collected from scratch in this work to support the proposed new task.

\vspace{0.5em}
\noindent\textbf{What (other) tasks could the dataset be used for?}
The dataset was created to support the CapOnImage task.  In addition, since each product image is accompanied by a product title sentence, it can be directly used for the Fashion Captioning \cite{yang2020facad} task. It may support a wider range of vision-and-language tasks as well. 

\subsection{Distribution}

\noindent\textbf{Will the dataset be distributed to third parties outside of the entity (e.g., company, institution, organization) on behalf of which the dataset was created?}
Yes. The dataset will be released publicly.

\vspace{0.5em}
\noindent\textbf{How will the dataset will be distributed (e.g., tarball on website, API, GitHub)?}
The dataset can be downloaded from Google Drive and Baidu Disk as a gzipped tar file.

\vspace{0.5em}
\noindent\textbf{When will the dataset be distributed?}
The dataset will be released upon the publication of this work.

\vspace{0.5em}
\noindent\textbf{Will the dataset be distributed under a copyright or other intellectual property (IP) license, and/or under applicable terms of use (ToU)?}
There will be no license. Users only need to fill in an agreement form regarding the dataset not to be used for commercial purposes and citation suggestions etc. 

\vspace{0.5em}
\noindent\textbf{Have any third parties imposed IP-based or other restrictions on the data associated with the instances?}
No. There are no fees or restrictions.

\subsection{Maintenance}

\noindent\textbf{Will the dataset be updated (e.g., to correct labeling errors, add new instances, delete instances)?}
Yes. The dataset will be updated for fair comparison with future works if there is any kind of changes.

\section{More Qualitative Results}
Figure~\ref{fig:demo1} illustrates more qualitative results of captioning on image through our \emph{full} model.
It is shown that our model can generate diverse and appropriate dense captions at different locations on the image.

\end{document}